\documentclass[10pt,twocolumn,letterpaper]{article}

\usepackage[pagenumbers]{cvpr} 

\newif\ifdraft
\drafttrue  
\ifdraft

    \newcommand{\francesc}[1]{\textcolor{orange}{#1}}
    \newcommand{\francescrmk}[1]{{\color{orange} {[\bf fmn: #1]}}}
\else
    \newcommand{\francesc}[1]{}
    \newcommand{\francescrmk}[1]{}
\fi

\usepackage{caption}

\renewcommand{\paragraph}[2][\ ]{\vspace{4pt}\noindent{\bf #2#1}}

\usepackage{adjustbox}

\usepackage{xcolor}

\newcommand{\ourdm}{PLACID}

\newcommand{\refmainfig}[1]{Fig.\textcolor{cvprblue}{~#1}}
\newcommand{\refmaintable}[1]{Tab.\textcolor{cvprblue}{~#1}}
\newcommand{\refmainsec}[1]{Sec.\textcolor{cvprblue}{~#1}}

\definecolor{cvprblue}{rgb}{0.21,0.49,0.74}
\usepackage[pagebackref,breaklinks,colorlinks,allcolors=cvprblue]{hyperref}

\usepackage{subfiles}
\usepackage{xr}

\def\paperID{8628} 
\def\confName{CVPR}
\def\confYear{2026}

\title{PLACID: Identity-Preserving Multi-Object Compositing\\ via Video Diffusion with Synthetic Trajectories}

\author{Gemma Canet Tarrés \hspace{5mm}Manel Baradad\hspace{5mm}Francesc Moreno-Noguer\hspace{5mm}Yumeng Li\\
Amazon\\
Barcelona\\
{\tt\small \{cangemma,mbaradad,cescmore,yumengll\}@amazon.es  }
}

\makeatletter
\renewenvironment{abstract}{
    \centerline{\large\bf Abstract}
    \vspace{-0.1em} 
   \vspace*{12pt}\noindent%
   \it\ignorespaces%
    
}{
    \par
}
\makeatother

\begin{document}

\twocolumn[{%
  \renewcommand\twocolumn[1][]{#1}%
  \maketitle
  \vspace{-18pt}
  \begin{center}
    \includegraphics[width=\linewidth]{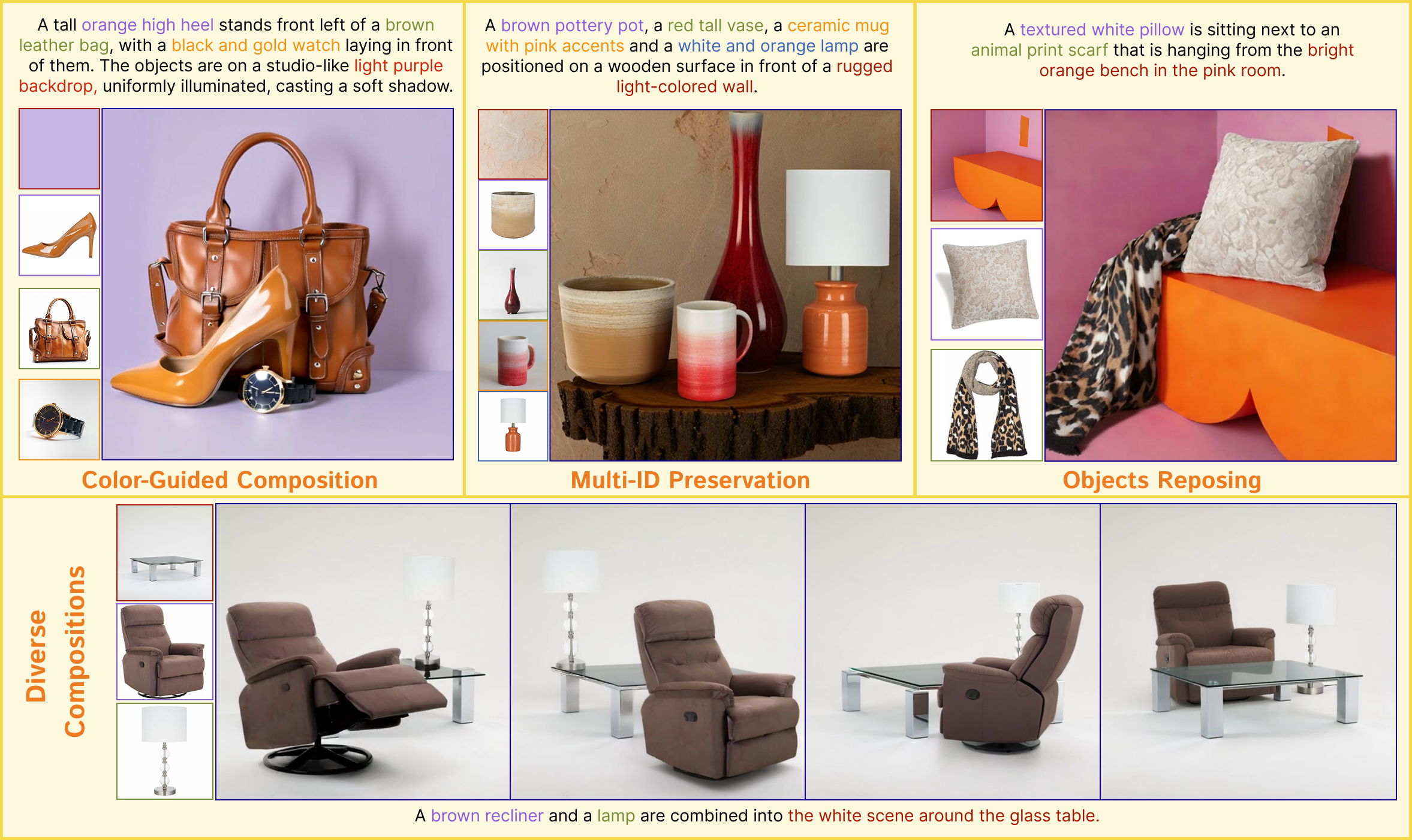}
    \captionof{figure}{%
        Our model enables simultaneous compositing of multiple objects into seamless natural scenes.
        Generation can optionally be guided by descriptive captions (\textit{top}), including specific color instructions (\textit{top left}).
        The model ensures all specified objects are present (\textit{top middle}) and allows them to interact with the given background through specific instructions (\textit{top right}) or creative compositions (\textit{bottom}).
    }
    \label{fig:teaser}
  \end{center}
  \vspace{5pt}
}]

\begin{abstract}
\vspace{-1em}
\noindent Recent advances in generative AI have dramatically improved photorealistic image synthesis, yet they fall short for studio-level multi-object compositing. This task demands simultaneous (i) near‑perfect preservation of each item’s identity, (ii) precise background and color fidelity, (iii) layout and design elements control, and (iv) complete, appealing displays showcasing all objects.
However, current state-of-the-art models often alter object details, omit or duplicate objects, and produce layouts with incorrect relative sizing or inconsistent item presentations.  
To bridge this gap, we introduce {\ourdm}, a framework that transforms a collection of object images into an appealing multi-object composite.
Our approach makes two main contributions. First, we leverage a pretrained image-to-video (I2V) diffusion model with text control to preserve objects consistency, identities, and background details by exploiting temporal priors from videos.
Second, we propose a novel data curation strategy that generates synthetic sequences where randomly placed objects smoothly move to their target positions. This synthetic data aligns with the video model’s temporal priors during training. 
At inference, objects initialized at random positions consistently converge into coherent layouts guided by text, with the final frame serving as the composite image. 
Extensive quantitative evaluations and user studies demonstrate that {\ourdm} surpasses state-of-the-art methods in multi-object compositing, achieving superior identity, background, and color preservation, with less omitted objects and visually appealing results.

\end{abstract}    
\vspace{-0.3em}
\section{Introduction}
\label{sec:intro}

In today's digital age, high-quality imagery plays a crucial role across various industries and applications. From e-commerce and marketing to graphic design, storytelling, and content creation, a single, well-crafted photograph can be the deciding factor between a user engaging or scrolling away, making visual content a critical driver of sales, brand identity, and user engagement across websites, social media, and digital catalogs. Creating compelling multi-object visuals, as shown in \cref{fig:teaser}, traditionally involves a complex chain of specialist steps: photographers shooting items under controlled lighting, designers isolating and re-assembling elements, adjusting lighting and shadows, and refining the result to meet exact brand guidelines or design standards. Each of these stages demands skilled professionals, specialized software, and considerable time, limiting creative possibilities and turning the workflow into a costly bottleneck when many items are involved. 

\begin{figure}[t]
    \centering    \includegraphics[width=\linewidth]{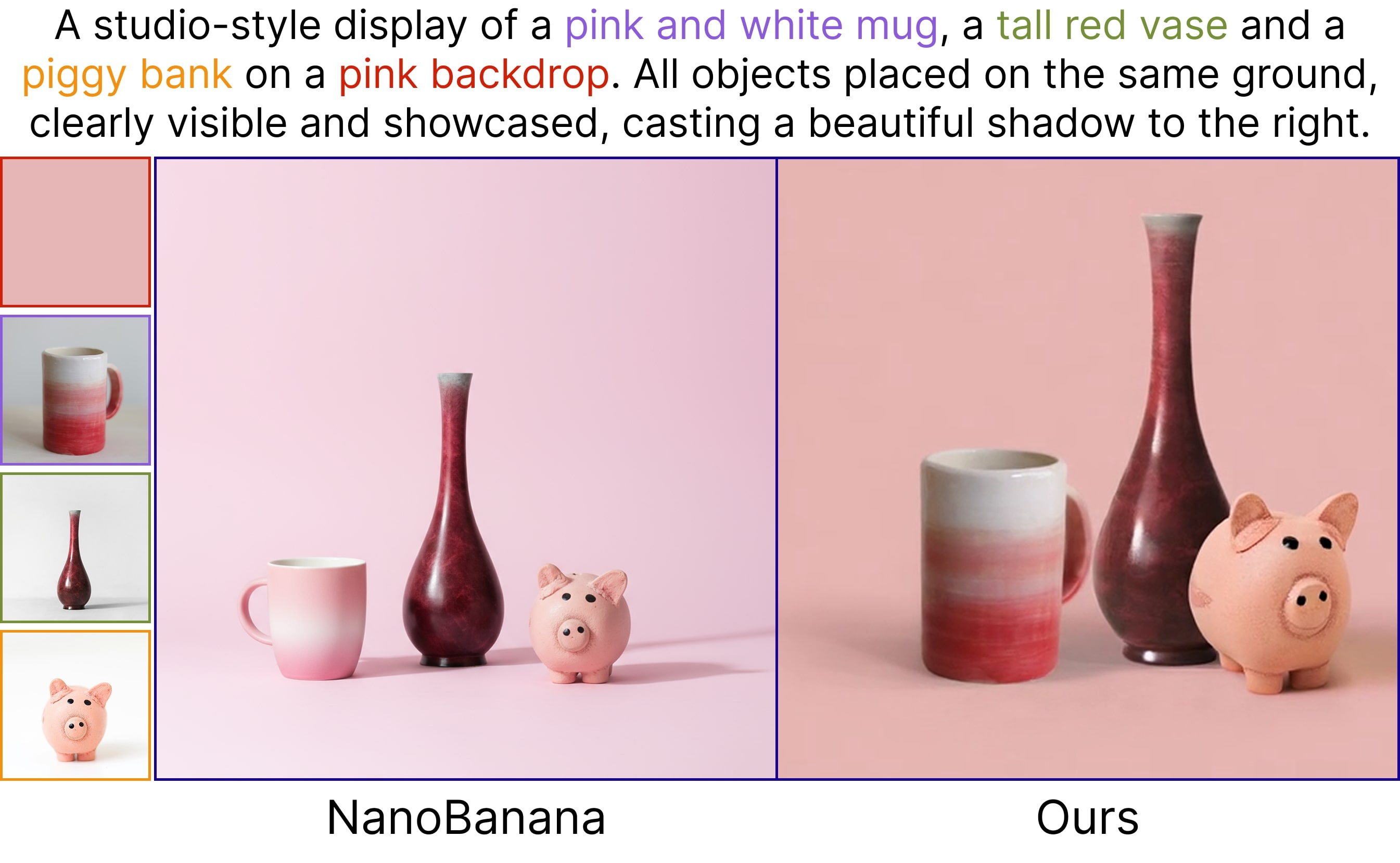}
    \vspace{-1em}
    \caption{Limitations of \cite{comanici2025gemini} in color accuracy and object consistency. Models often (i) alter colors, and (ii) omit, duplicate, or modify objects when combining multiple items.} 
    \label{fig:intro}
    \vspace{-1.4em}
\end{figure}

Recent advances in diffusion-based generative models \cite{seedream2025seedream,wang2025wan,ju2025editverse,labs2025flux} have dramatically improved the realism of synthetic images and videos. However, these models, primarily trained on generic ``internet-style" data, struggle with the specific demands of studio-quality multi-object photography, as shown in \cref{fig:intro}. 
This task uniquely combines the challenges of multi-object compositing with the demanding requirements of professional studio-based imagery, leading to four essential requirements:
\textbf{(i) High identity preservation:} maintaining high fidelity to each object's identity, including specific color gradients, textures, and shapes, is crucial for effective ready-to-use images. 
\textbf{(ii) Fine-grained background and color fidelity:} in photography and visual design, 
colors are carefully chosen to evoke specific emotions or convey brand identity; even subtle shifts may confuse viewers or alter perception. Accurate preservation of specific background and design colors is therefore essential for producing images ready for professional use. 
\textbf{(iii) Design controllability:} while automatic generations can be used to boost creativity, users may sometimes desire some control over aspects such as objects distribution, lighting, and additional elements. Such flexibility enables refined designs that can effectively convey targeted messages.
\textbf{(iv) Complete, appealing displays:} The composite image must include all intended items, without duplicates or omissions, properly scaled, coherently posed, and arranged in a visually pleasing composition. 

Current diffusion models cannot consistently meet all these criteria simultaneously, falling short in reliably replacing human studio work for this task. This limitation highlights the need for a specialized and user-friendly framework. In this paper, we address these challenges with \textbf{PLACID} (PLACing objects with ID preservation using Diffusion), our proposed model that leverages pre-trained image-to-video (I2V) diffusion models with text guidance for multi-object compositing. Compared to image-to-image (I2I) models, I2V models possess stronger priors on object interactions and reposing. This inherent knowledge enables {\ourdm} to create more realistic layouts while preserving object identities.
However, a critical challenge emerges: typical video training data rarely features isolated inanimate objects moving independently, making it unsuitable for our task.
To overcome this limitation, we introduce a novel training data generation technique. Simply interpolating between an initial randomly assembled frame and a ground truth final frame in a naive way is insufficient, as it lacks temporal coherence and would negate the temporal priors of videos that we aim to exploit. Instead, {\ourdm} synthesizes short videos where objects follow smooth, physically plausible trajectories from initial random locations to desired final positions, aligned with provided text input. This temporal scaffold helps preserve object identity during movement, prevents object erasure or duplication, and provides a dynamic progression chain that regularizes the generation process. The last frame of this synthetic video becomes the desired composite image, as shown in \cref{fig:teaser}. By generating appropriate training data featuring objects moving independently in a physically plausible manner, we enable {\ourdm} to effectively leverage the temporal priors of video models while adapting to the specific requirements of object compositing and adhering to text-based instructions.

In summary, our contributions are as follows:
\noindent\textbf{(i) A Novel Model for Ready-To-Use Multi-Object Compositing.} We introduce {\ourdm}, a text- and image-guided model designed primarily for high-fidelity, multi-object compositing with fine-grained background and color preservation. While optimized for compositing objects on given backgrounds, {\ourdm} is versatile: it can add, rearrange, or replace objects in existing images, incorporate text-described elements, and generate text-aligned scenes. 

\noindent\textbf{(ii) Training Data Generation Pipeline.} We design a synthetic data pipeline to produce short videos where objects move in a linear trajectory from random initial positions to final composite layouts. This approach aligns with the pretrained video model's temporal priors, enabling us to leverage its capabilities for multi-object compositing scenarios. 

\noindent\textbf{(iii) Comprehensive Evaluation.} We conduct extensive quantitative and qualitative analyses to demonstrate our model's effectiveness in addressing key challenges: identity preservation, color accuracy, text alignment, and compositional quality. These experiments reveal our model's superiority over state-of-the-art approaches in preserving object identities, background details, and fine-grained colors, while ensuring coherent composite results. 


\vspace{-0.2em}
\section{Related Work}
\label{sec:soa}

\textbf{Generative Object Compositing.} Object compositing is a crucial task in many editing and content creation pipelines, aiming to seamlessly insert one or more foreground objects into a background scene. Traditionally approached via harmonization \cite{guerreiro2023pct,jiang2021ssh} and blending techniques \cite{zhang2021deep,perez2023poisson}, current approaches predominantly adapt text-to-image (T2I) diffusion models for this task.
Methods like \cite{song2023objectstitch,yang2023paint} successfully adapt to the task using CLIP-based adapters but struggle with identity preservation. 
More recent models, such as AnyDoor \cite{chen2024anydoor} and IMPRINT \cite{song2024imprint}, achieve improved identity preservation without compromising natural object integration into the background. However, these models typically accept only one object as input and require a bounding box or mask to specify the desired object location. 
\cite{canet2024thinking} proposed the first object compositing model capable of inserting an object without a location cue, but it doesn't support multiple objects or textual input for guiding the generation. Conversely, \cite{tarres2025multitwine} allows users to simultaneously composite multiple objects  with text guidance but requires additional bounding box inputs. Thus, none of these models can be directly used with our same set of inputs: multiple objects, a background image, and a descriptive caption. 

\vspace{1mm}\noindent\textbf{Subject-Driven Generation.} Subject-driven models generate new images integrating specific subjects based on visual examples. Early approaches like DreamBooth \cite{ruiz2023dreambooth}, CustomDiffusion \cite{kumari2023multi}, and Textual Inversion \cite{gal2022image} pioneered personalized text-to-image generation through fine-tuning, but lacked scalability. Subsequent methods like \cite{shi2024instantbooth,chen2023subject} improved efficiency with real-time subject incorporation. BLIP-Diffusion \cite{li2023blip} and IP-Adapter \cite{ye2023ip} introduced zero-shot techniques, though they struggled with text alignment and multi-subject guidance. 
Recent advancements include MS-Diffusion \cite{wang2024ms} with improved multi-subject conditioning by focusing on regrounding and cross-attention, and UNO \cite{wu2025uno}, offering improved quality via curated data. OmniGen \cite{xiao2025omnigen} leverages multi-task training data for improving generation, and DSD \cite{cai2025dsd} leverages video priors for better identity preservation. Despite progress, challenges persist in maintaining consistent subject appearance across multiple objects and balancing identity preservation with text alignment. When a background scene is provided, it is generally processed similarly to object references, resulting in semantically consistent but detail-altered scenes, which complicates its use for the task of object compositing.

\vspace{1mm}\noindent\textbf{Image Editing Models.}    
The emergence of instruction-based methods \cite{brooks2023instructpix2pix,zhang2023magicbrush,hertz2022prompt} advanced usability by enabling direct natural-language editing without per-image fine-tuning. 
Recent models such as Qwen Image Edit \cite{wu2025qwen} and NanoBanana \cite{comanici2025gemini} improve user control and editing fidelity, enabling fine-grained and consistent language-guided edits.
VACE \cite{jiang2025vace} extends the paradigm to video, accepting text, images, masks, and control signals to support reference-to-video and masked editing workflows. 
Some recent works \cite{yu2025objectmover,lin2025realgeneral,wu2025chronoedit} leverage I2V models for image editing, focusing on single-object or style edits. In contrast, {\ourdm} exploits video priors for multi-object compositing with consistent color-based or photorealistic backgrounds.

\section{Methodology}
\label{sec:methodology}


Inspired by video models' strong priors on identity preservation, object interactions, and reposing, we propose adapting a pre-trained text-guided I2V model \cite{wang2025wan} for multi-object compositing. The pipeline and architecture are detailed in \cref{sec:model}, followed by the training strategy in \cref{sec:training-strategy}, and training data curation process in \cref{sec:data}.


\subsection{Model Architecture}
\label{sec:model}

To generate a composite image showcasing $N$ object, our framework accepts three user‑provided inputs: (i) an optional background image $B$; (ii) an unsegmented photograph $I_i$ for each item $i \in 1..N$; and (iii) a free‑form caption $c$ that optionally describes the desired output, including layout, lighting, props, or any additional detail. Initially, the $N$ object images are placed at random positions on the background $B$ (or on a white canvas when no background image is supplied), producing a naive composite that we denote $F_{1}$. The core of our system then synthesises a K‑frame video $V = \{F_{1},..,F_{K}\}$ that smoothly transforms $F_1$ into a polished multi-object scene $F_K$ as described in $c$. If no background scene $B$ is provided, the white canvas in $F_1$ progressively fades into the final scene described in $c$. The last frame $F_K$ serves as the final composite image.


\begin{figure}
    \centering
    \includegraphics[width=\linewidth]{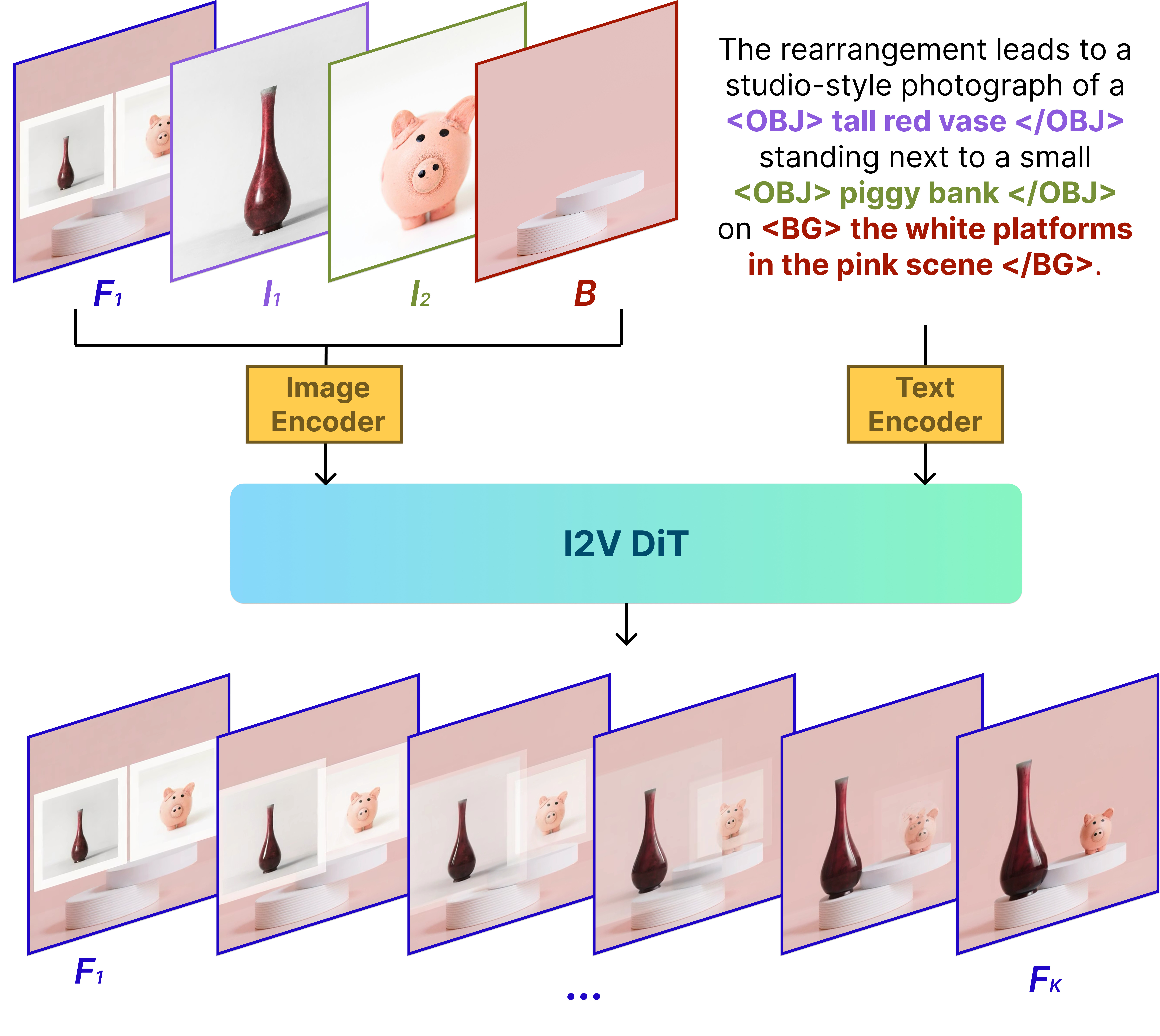}
    \caption{Model Architecture. Our model conditions a DiT video model on images and text. The visual inputs, encoded via \cite{radford2021learningCLIP}, include: (i) first frame $F_1$ (a random assembly of unprocessed object images), (ii) individual object images $I_{1..N}$, and (iii) an optional background $B$. A caption $c$ describing the desired composition is encoded via \cite{raffel2020exploringT5}. Image and text encodings are fed to the DiT through separate cross-attention mechanisms. The model flexibly handles varying numbers of objects, with or without $B$.} 
    \label{fig:diagram}
\vspace{-1.4em}
\end{figure}


As illustrated in \cref{fig:diagram}, our architecture builds upon an image‑to‑video diffusion transformer (DiT)~\cite{wang2025wan} with text guidance. The original DiT conditions the generation on two inputs  via  separate cross‑attention modules: (i) the first frame and (ii) a caption. 
We introduce two key modifications to adapt this pipeline for multi-object compositing.

\vspace{1mm}\noindent\textbf{(i) High‑Resolution Object‑Image Conditioning.} In the original DiT only down‑sampled object images in $F_1$ are fed to the model, which can inevitably discard their fine details or lead to identities entanglement. We therefore alter the visual guidance of the model by concatenating the original, full‑resolution object images $I_{1..N}$ to the naive composite $F_1$. We also concatenate the full background image $B$, if provided, to preserve details potentially occluded by object images in $F_1$. These images are individually encoded via CLIP \cite{radford2021learningCLIP}, adapted as in \cite{wang2025wan} and fed via cross-attention to condition video generation. This enhanced visual guidance helps preserve background details throughout the diffusion process and prevents blurring or mixing of object details that occurs when using only the $F_1$ composite as guidance. 

\vspace{1mm}\noindent\textbf{(ii) Object‑Aware Textual Tokens.} To unambiguously reference each object and the background in the caption, we introduce four special tokens: $<\textit{OBJ}>, </\textit{OBJ}>, <\textit{BG}>, </\textit{BG}>$. These tokens delimit descriptions of specific objects and the overall background, matching the order of provided images. During textual conditioning cross-attention, these tokens guide the association of each enclosed fragment with its corresponding visual information, while remaining descriptions complete the image. 

\subsection{Training Strategy}\label{sec:training-strategy}
The pretrained I2V DiT architecture \cite{wang2025wan} inherently accepts our additional textual and visual conditioning inputs. Thus,
we fine-tune the base architecture using a lightweight LoRA adapter \cite{hu2022lora}, which learns to fuse the high‑fidelity visual cues with object‑specific textual tokens without overwriting original weights. 
The training is done by using Flow Matching \cite{lipmanflow} and minimizing the following loss function:
\vspace{-0.5em}\begin{equation}
    \mathcal{L} = \mathbb{E}_{V_{0},\,V,\,I_c, \,c,\,t}
\bigl\|\,u(x_{t},\,I_c, \,c,\,t;\,\theta) - w_{t}\,\bigr\|^{2}\,,
\end{equation}
where $V_0 \sim \mathcal{N}(0,1)$ is random noise, $t \in [0,1]$ is a sampled timestep, $V_t = tV + (1-t)V_0$ is the noisy input obtained by linearly interpolating the noise $V_0$ with the ground-truth video $V$, $\theta$ are the model weights, and $u(x_{t},\,I_c,\,c,\,t;\,\theta)$ denotes the output velocity predicted by the model, guided by the visual inputs $I_c = [F_1, I_1, ..., I_N]$, and textual input $c$. The model is designed to predict the velocity $w_t = \frac{dV_t}{dt} = V_t-V_0$. 

As detailed in \cref{sec:data}, our curated training data uses synthetic trajectories to ensure temporal and spatial consistency of objects, making all frames (initial, intermediate, and final) meaningful. Consequently, the loss targets the entire video. However, for the subset of data where meaningful intermediate frames can't be guaranteed (\cref{sec:data} (ii)), only the last frame is used in loss computation.







\subsection{Training Data Generation} 
\label{sec:data}

Our I2V model generates smooth transitions, transforming a naive composition of randomly scattered, unsegmented object images on an optional background into a natural-looking, seamless multi-object scene, guided by text. The final frame serves as the desired multi-object composite.
Training requires paired examples containing: (i) Object images $I_{1\ldots N}$, (ii) Optional background $B$, (iii) Ground-truth multi-object scene $F_K$, (iv) Video animating items from initial to final positions $V = {F_1\ldots F_K}$, (v) Caption $c$ describing target composition and required transformations.

\begin{figure}
    \centering
    \includegraphics[width=\linewidth]{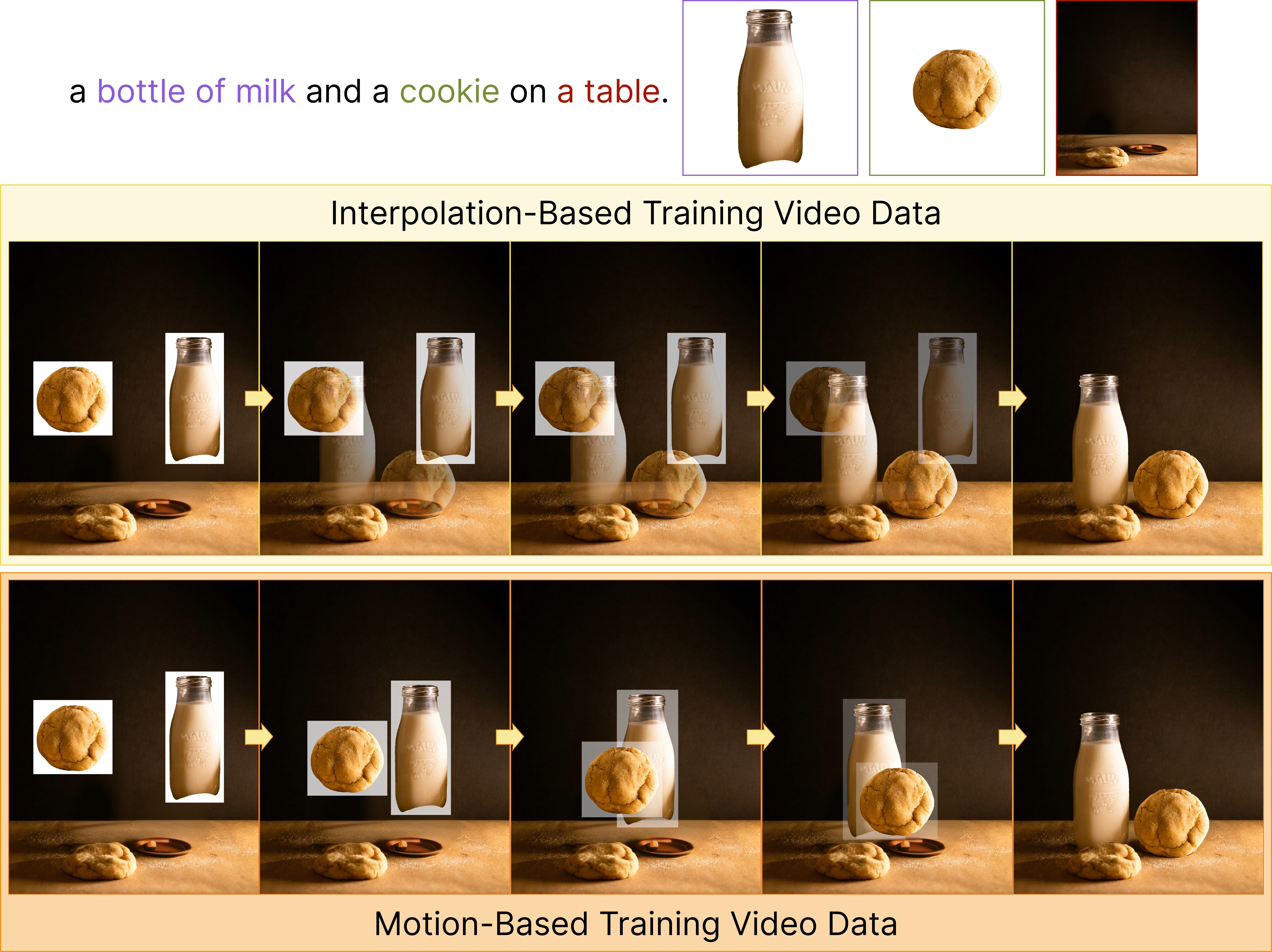}
    \caption{Training Data Generation. \textbf{Top:} Naive way of generating training video data, by simply interpolating the first and last frame. \textbf{Bottom:} Proposed method of generating motion-based temporal consistent video data, in which objects follow synthetic trajectories from initial to final position. Final frame from \cite{unsplash}.}
    \label{fig:trainingdata}
    \vspace{-1.7em}
\end{figure}

Existing video datasets \cite{wanginternvid,nanopenvid,chen2024panda,qin2024instructvid2vid,zi2025senorita,hu2024vivid}, including object-focused ones \cite{yuan2025opens2v}, predominantly feature either objects in dynamic scenes with human interaction or naturally moving items (\eg, vehicles, animated toys). However, they rarely show inanimate objects moving independently, as they don't exhibit autonomous motion in real-world scenarios. These datasets are thus unsuitable for our task, which often involves static, inanimate objects. To address this challenge, we propose a novel approach to synthesize the required training video data from still images showcasing multiple objects. 
Given a target image with multiple objects $F_K$ and an initial naive composition $F_1$, the simplest approach to generate the required video animation would be progressive interpolation between initial and final frames (\cref{fig:trainingdata} top). However, this produces intermediate frames with half-faded objects in both positions, lacking motion consistency and obscuring object correspondence. Such data would negate the benefits of using a video model, including priors on identity consistency and prevention of arbitrary object appearance or disappearance. Instead, we propose animations that move objects from initial to final positions using linear synthetic trajectories (\cref{fig:trainingdata} bottom). This method maintains temporal and spatial consistency, maximizing video model advantages and discouraging identity changes, as well as object mixing, dropping, or duplication. We obtain data from the following three modalities, with more details provided in SupMat.

\vspace{1mm}\noindent\textbf{(i) Professional Multi-Object Images.} We utilize two sources of professional photography: (a) In-the-Wild Images and (b) Manual Designs. \textbf{(a) In-the-Wild Images:} We process `Product Photography' and `Flat Lay' images from Unsplash \cite{unsplash} with their captions. Using GroundingDINO \cite{liu2024grounding} and SAM \cite{kirillov2023segment} as in \cite{ren2024grounded}, we identify, filter and refine object detections. We alternate between inpainting \cite{yu2023inpaint} a clean background scene and randomly generating a new one. 
Segmented objects are pasted onto white boxes, emulating test-time object images, and scattered on the background or a white canvas to create $F_1$. Objects follow synthetic trajectories to their final position in $F_K$, white boxes fade out, and the scene is completed ( \cref{fig:trainingdata} (bottom)). \textbf{(b) Manual Designs:} To address limitations of in-the-wild images (occluded, incomplete objects), we ask designers to curate $\sim 400$ compositions of 2-5 objects from separate high-resolution images. We use the same pipeline to construct $F_1$ and synthesize the video. Captions combine template prompts with object and background descriptions generated by \cite{bai2025qwen2}. However, neither source offers drastic reposing or relighting of objects, and descriptive captions lack diversity.

\vspace{1mm}\noindent\textbf{(ii) Subject-Driven Generation Paired Dataset.} To enable more diverse text descriptions, allow object reposing, and ensure more natural and realistic scenes, we use a filtered subset of the Subject-200k dataset \cite{tan2024ominicontrol}. Each pair includes a white-background object image ($F_1$) and the same item in context ($F_K$), with a descriptive caption. We filter pairs using GroundingDINO \cite{liu2024grounding}, with high-confidence bounding boxes ensuring consistent object identity. For significant reposes, we interpolate the object throughout the K-frame video and adapt the training objective (\cref{sec:training-strategy}) to focus solely on the last frame for this data source.  This approach enhances our model's ability to maintain text alignment and generate contextually appropriate object placements.

\vspace{1mm}\noindent\textbf{(iii) Synthetic Side-by-side Compositions.} To enhance our model's ability to cohesively relight and rescale objects, we create a set of animations placing two objects side-by-side in the final frame. Inspired by \cite{collins2022abo}, we gather $~14k$ synthetic 3D renders of objects with known dimensions. For $F_K$, we apply consistent lighting, cluster all objects into three size groups using K-Means \cite{mcqueen1967somekmeans}, select pairs from the same group, and place them side-by-side, scaled relative to their real-world dimensions on a random background. For $F_1$, we scatter randomly relighted images of the same objects on the background. The K-frame video animates objects from initial to final positions with progressive relighting and transformation following synthetic trajectories. Captions combine templates with object descriptions generated via \cite{bai2025qwen2}.

The three data types above yield about $50K$ diverse, annotated (background, object images, target composition, transition video, caption) tuples, enabling our model to transform naive composites into high-fidelity, caption-driven displays. We enhance this data with various augmentations: \textbf{(a) Object Augmentations.} During training, objects are randomly rescaled, rotated, and warped to discourage direct copying. \textbf{(b) Background Augmentations.} We alternate using lifestyle images, plain RGB colors, and generated backgrounds with simple primitives, improving adaptability to various background types. \textbf{(c) Caption Augmentations.} To enhance text alignment and improve robustness to caption diversity, we introduce object replacement and alternately treat some objects as part of the background or as design elements. These augmentations extend the model's usability, adaptability, visual quality, and diversity. More augmentation details can be found in SupMat.

\begin{figure*}
    \centering
    \includegraphics[width=\linewidth]{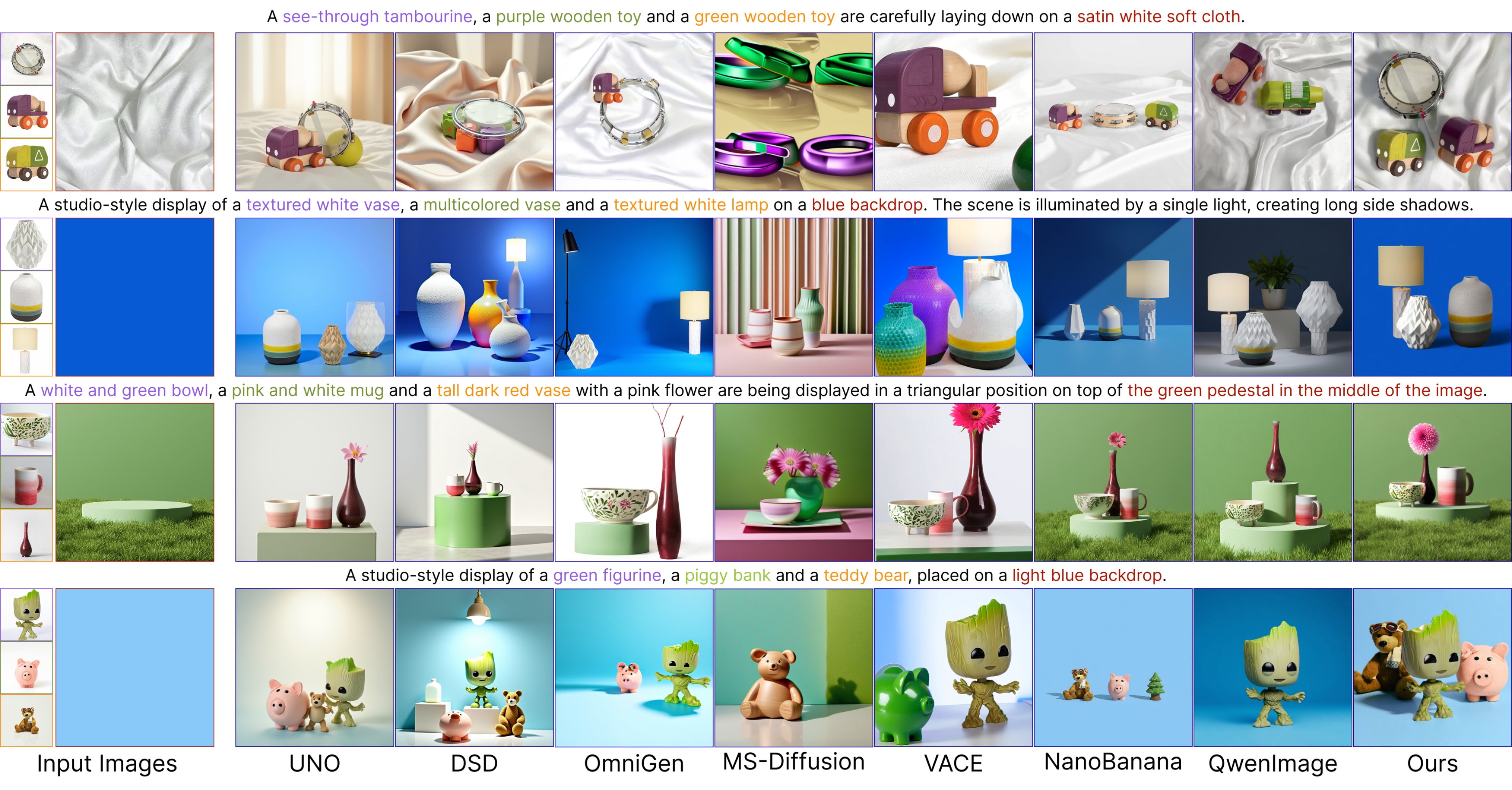}
    \vspace{-0.4em}\caption{Comparison to State of the Art. We compare our multi-object compositing model to VACE \cite{jiang2025vace}, UNO \cite{wu2025uno}, DSD \cite{cai2025dsd}, OmniGen \cite{xiao2025omnigen}, MS-Diffusion \cite{wang2024ms}, NanoBanana \cite{comanici2025gemini} and Qwen-Image-Edit \cite{wu2025qwen}.}
    \label{fig:soa}
    \vspace{-1.7em}
\end{figure*}

\vspace{-0.3em}
\section{Experiments}
\label{sec:experiments}
\vspace{-0.5em}

\paragraph{Evaluation Dataset.} Our model is specifically engineered to generate ready-to-use images showcasing multiple objects. To assess its performance, we combine product images from the Amazon Berkeley Objects Dataset (ABO) \cite{collins2022abo} with objects from DreamBench++ \cite{peng2024dreambench}, ensuring a mix of diverse items and objects with challenging identities. We randomly group these into 122 sets of 1-7 objects, pairing each set with either a plain-color canvas or an image from Unsplash \cite{unsplash} as background. Captions are generated in various styles and lengths, with some deliberately describing additional design elements or layouts to rigorously test text-to-image alignment. This evaluation protocol creates a realistic benchmark reflecting the challenges of real-world multi-object image generation, incorporating varying image resolutions, different levels of text control, and diverse background types. More details in SupMat. 

\vspace{-0.2em}
\vspace{1mm}\noindent\textbf{Evaluation Metrics.} To comprehensively evaluate our model and compare it with other works, we employ a range of metrics addressing five key aspects: (i) identity preservation, (ii) background preservation, (iii) text alignment, (iv) color fidelity, and (v) image quality and coherence.
First, we use GroundingDINO \cite{liu2024grounding} to detect each object mentioned in the provided text caption. Objects not detected are considered `missing', and we report the percentage of missing objects over the total expected objects (\textbf{Missing}) as an evaluation metric. For assessing identity preservation, when objects are successfully detected, we crop their images and compare them to the reference images using CLIP-Score \cite{hessel2021clipscore} (\textbf{CLIP-I}) and DINO-Score \cite{oquab2024dinov2} (\textbf{DINO}). For scenes with multiple objects, we average these scores across all objects. 
We assess text alignment using CLIP-Score between generated image and provided caption (\textbf{CLIP-T}). For cases requiring specific background colors, we quantify color fidelity using the average Chamfer Distance (\textbf{Chamfer}) between segmented background and provided color. For the remaining background scenes, their preservation is evaluated by segmenting the background and computing Mean Squared Error (\textbf{MSE-BG}). Both background metrics are normalized by number of pixels in segmented background. 


Recognizing that quantitative metrics often miss nuanced aspects of image quality and coherence (\eg, identity preservation scores can be maximized by copy-pasting reference images, yielding incoherent scenes), we complement our evaluation with two user studies assessing (i) \textit{identity preservation} across composited objects, and (ii) \textit{overall composition quality}, including scene coherence and seamless integration. In each study, users compare our model’s output with others side by side and indicate a preference or select neutral if the quality is similar. By combining quantitative metrics and user studies, we provide a comprehensive evaluation addressing both objective measures and subjective human perception in image generation.
\vspace{-0.2em}
\vspace{1mm}\noindent\textbf{Training/Testing Details.} {\ourdm} is trained end-to-end on 8 H100 GPUs using Adam optimizer with a learning rate of $2 \times10^{-5}$ for 119k iterations. We use 9-frame sequences for training ($K=9$), balancing object trajectory understanding with computational efficiency, while allowing longer sequences at test time. For evaluation, we generate 33-frame videos. Object images are randomly scattered to form $F_1$ during both training and evaluation. The effect of varying $F_1$ and number of frames is explored in SupMat. 


\vspace{-0.4em}

\subsection{Comparison to State of the Art}
\label{sec:soacomparison}
\vspace{-0.1em}

{\ourdm} is designed for multi-object compositing. While several state-of-the-art methods focus only on object compositing, they are limited by single-object operation \cite{canet2024thinking,song2024imprint}, require explicit bounding boxes \cite{chen2024anydoor,song2023objectstitch,yang2023paint}, or lack of open-source code \cite{tarres2025multitwine}. We compare {\ourdm} to a new generation of multi-purpose models that handle the same inputs: (i) multiple object images, (ii) a background image, and (iii) a text prompt. These include models for multi-subject guided image generation (UNO \cite{wu2025uno}, DSD \cite{cai2025dsd}, OmniGen \cite{xiao2025omnigen}, MS-Diffusion \cite{wang2024ms}) and image/video editing (VACE \cite{jiang2025vace}, NanoBanana \cite{comanici2025gemini}, Qwen Image Edit 2509 \cite{wu2025qwen}).

As reported in \cref{tab:soa,fig:soa}, {\ourdm} outperforms existing approaches across multiple metrics. It significantly improves color preservation and background fidelity, evidenced by reduced Chamfer Distance and background MSE scores. {\ourdm} maintains background colors while incorporating shadows and lighting gradients (\cref{fig:soa} rows 2, 4, \cref{fig:teaser} top-left), and preserves photorealistic backgrounds without altering camera angles (\cref{fig:soa} rows 1, 3). The temporal coherence in our motion-based training data and the use of a video model result in fewer missing objects. Text alignment is comparable to state-of-the-art models, but {\ourdm} prioritizes visual over strict text alignment when ambiguous (\eg, `green figurine' in \cref{fig:soa} last row). {\ourdm} optimizes for object identity preservation while creating natural, appealing compositions. This approach may result in slight object overlaps or reposing, with required new view synthesis. While it ensures visual coherence, it might yield lower CLIP-I or DINO scores than other models. To capture the nuanced balance between identity preservation and compositional quality and coherence, we rely on user studies that complement quantitative metrics, providing a more comprehensive evaluation. 

\begin{table}[t!]

\centering
\begin{adjustbox}{width=\linewidth}

\begin{tabular}{lcccccc}
\toprule
\textbf{Method}                            & \textbf{CLIP-I$\uparrow$} & \textbf{DINO$\uparrow$} & \textbf{CLIP-T$\uparrow$} &  \textbf{MSE-BG$\downarrow$} & \textbf{Chamfer $\downarrow$} & \textbf{Missing  $\downarrow$} \\

\midrule \multicolumn{6}{c}{\textit{Multi-Subject Guided Image Generation}}
\\ \midrule
UNO       &  0.696 & 0.450          &      0.346                                                                                              &  0.062    &   14.733          &   0.099 \\
DSD       &  0.650 & 0.362          &   \textbf{0.347}                                                                                                 &   0.083   &       11.886        &   0.102   \\
OmniGen       & \textbf{0.724}  & \textbf{0.478}         &       0.337  &   0.119         &     15.120 & 0.128  \\
MS-Diffusion       &  0.574 & 0.245          &   0.314                                                                                                 &   0.166   &    16.322         &  0.071 \\
\midrule \multicolumn{6}{c}{\textit{Image and Video Editing Models}}
\\ \midrule
VACE       &  0.689 & 0.439          &  0.343                                                                                                  &  0.096    &     9.948         &   0.096    \\
NanoBanana      &  0.662 & 0.390          &      0.344                                                                                              &     0.029 &     13.146        & 0.138        \\

Qwen     &  0.625 & 0.308          &      0.317                                                                                              &   0.097   &   49.317        & 0.115      \\
 \midrule

Ours     &  0.705 & 0.440          &            0.336                                                                                        &   \textbf{0.019}   &    \textbf{4.641}        &    \textbf{0.044}       \\

\bottomrule
\end{tabular}
\end{adjustbox}

\caption{Quantitative comparison to SOTA Models. Metrics: \textit{Missing} (percentage of objects missing in generated output), \textit{CLIP-I} and \textit{DINO} (object identity preservation), \textit{CLIP-T} (text alignment), \textit{Chamfer} (background color fidelity), and \textit{MSE-BG} (photorealistic background faithfulness).}
\label{tab:soa}
\vspace{-1.4em}
\end{table}

\begin{figure}
    \centering
    \includegraphics[width=\linewidth]{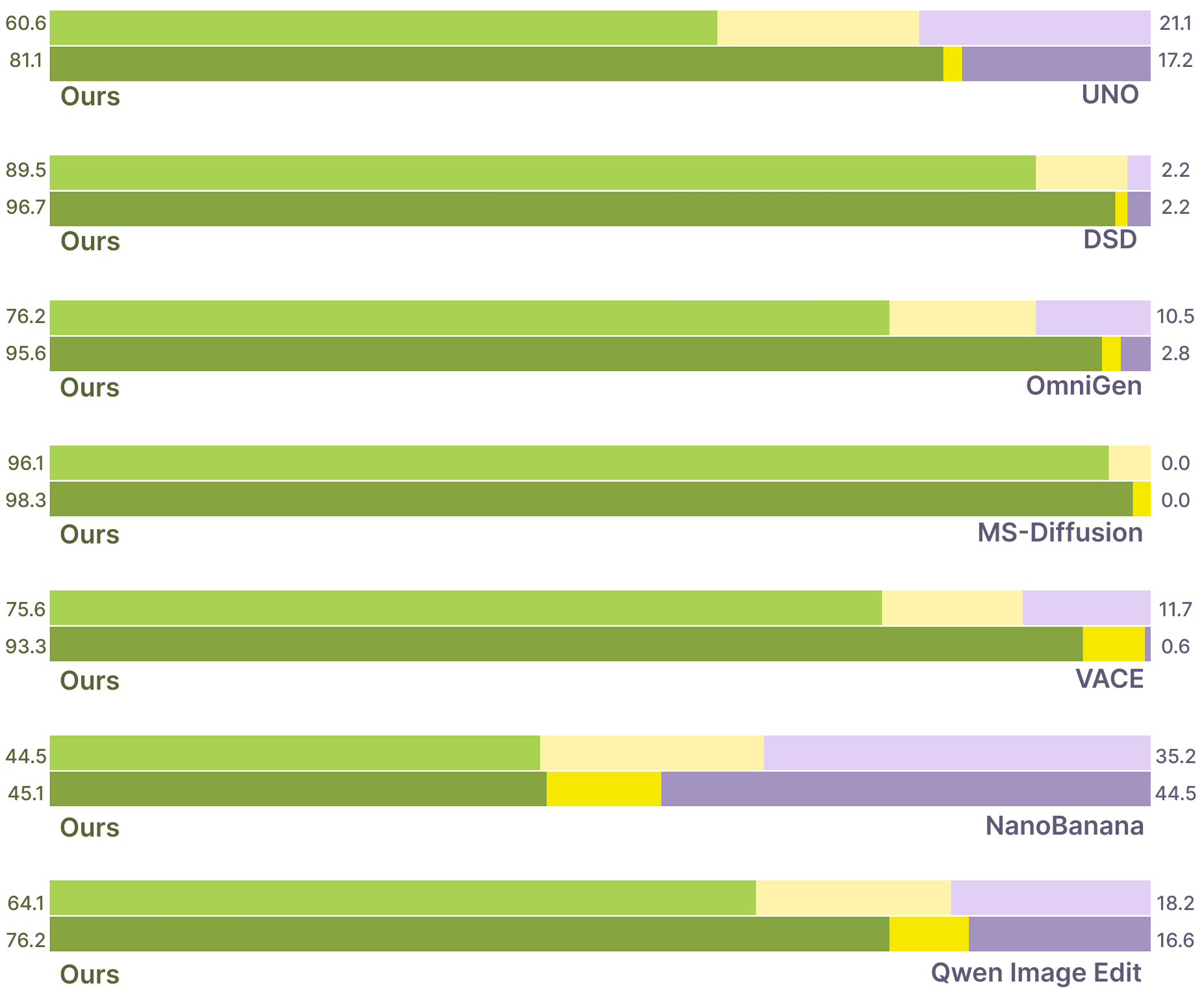}
    \caption{User study comparing {\ourdm} with SOTA models. Two user studies evaluate (i) identity preservation and (ii) overall visual preference. \textbf{Top:} Identity Preservation Preference. Light green: users preferred our model; light purple: users preferred the compared work; light yellow: no preference. \textbf{Bottom:} Overall Preference. Dark green: users preferred our model; dark purple: users preferred the compared work; dark yellow: no preference.}
    \label{fig:userstudies}
    \vspace{-1.4em}
\end{figure}

\vspace{1mm}\noindent\textbf{User Studies.} 
We conduct two user studies showing a total of 1265 side-by-side comparisons of our model and each of the other works to eight external users. Each user is asked to choose their preferred option in terms of identity preservation and overall quality, with an option to stay neutral. As shown in \cref{fig:userstudies}, our method significantly outperformed all open-source alternatives in both studies. Although the margin against the closed-source NanoBanana was smaller, our method still achieved a slight advantage in user preference.

\vspace{-0.4em}
\subsection{Ablation Studies}
\label{sec:ablations}
\vspace{-0.1em}

\noindent\textbf{Training Strategy.}
Ablation studies on our training pipeline are shown in \cref{tab:ablations} (top), starting with the base model \cite{wang2025wan}. The base model, when applied directly, struggles with our task due to distribution mismatch, minimally transforming objects while overly altering backgrounds. Fine-tuning on our data improves background and color preservation, but identity preservation remains challenging. Adding background and object images as extra inputs enhances identity preservation and reduces Chamfer Distance, but text and identity alignment remain suboptimal. Introducing tokens relating text to images improves performance across most metrics. Finally, using only last frame for loss computation with Subject-200k data reduces training noise, resulting in fewer omitted objects, improved text alignment and better photorealistic background preservation.

\paragraph{Training Data.} The bottom part of \cref{tab:ablations} presents ablation studies on different sources of training data. In-the-Wild data helps create complete scenes with well-preserved backgrounds and identities, but leads to copy-pasting behavior and limited text alignment. The professionally created manual set and side-by-side compositions provide improved fine-grained color preservation, but struggle with preserving photorealistic background scenes. Training on Subject-200k \cite{chen2023subject}, designed for subject-driven image generation, results in excellent text-image alignment, though at the cost of specific background and identity preservation. By combining all these sources, our model leverages their complementary benefits, achieving the desired balance between text alignment, background retention, and identity preservation. Visual comparison provided in SupMat.

\begin{table}[t!]

\centering
\begin{adjustbox}{width=\linewidth}

\begin{tabular}{lcccccc}
\toprule
\textbf{Method}                            & \textbf{CLIP-I$\uparrow$} & \textbf{DINO$\uparrow$} & \textbf{CLIP-T$\uparrow$} &  \textbf{MSE-BG$\downarrow$} & \textbf{Chamfer$\downarrow$} & \textbf{Missing$\downarrow$} \\
\midrule \multicolumn{6}{c}{\textit{Pipeline and Training Ablation}}
\\ \midrule
Wan 2.1       &  \textbf{0.711} &      0.446                                                                                               &   0.333   &     0.047       &      7.746 &  0.048          \\
\cmidrule{1-7}

\ \ + \{FT\} &    0.691 & 0.415        &   0.331                                                                                               &   0.042     &  5.754 & 0.045     \\
\cmidrule{1-7}
\ \ + \{FT, $I_c$\}  & 0.698 & 0.440     & 0.329 &   0.040                                                                                                               &   \textbf{4.138}  & \textbf{0.042}                     \\
\cmidrule{1-7}
\ \ + \{FT, $I_c$, TOK\}  & 0.703 & \textbf{0.447}     & 0.331   &    0.033                                                                                                             & 4.555  &      0.051             \\
\cmidrule{1-7}
\cmidrule{1-7}
Ours      &  0.705 & 0.440          &            \textbf{0.336}                                                                                        &   \textbf{0.019}   &    4.641      & 0.044            \\
\midrule \multicolumn{6}{c}{\textit{Training Data Sources Ablation}} \\
\midrule
In-the-Wild  & \textbf{0.721} & \textbf{0.465}     &  0.330  &  0.023                                                                                                             & 4.865  &   0.058                 \\
\cmidrule{1-7}
Manual Data  & 0.693 & 0.432     &  0.332  &  0.042                                                                                                             & 4.358  &   0.054                 \\ 
\cmidrule{1-7}
Subject-200k  & 0.683 & 0.373     &  \textbf{0.343}  &  0.129                                                                                                             & 9.882  &     0.048              \\
\cmidrule{1-7}
Side-by-side & 0.691 & 0.421     &  0.327  &     0.032                                                                                                          & \textbf{3.609}  &  0.045                  \\

\cmidrule{1-7}
\cmidrule{1-7}
All Sources     &  0.705 & 0.440          &            0.336                                                                                        &   \textbf{0.019}   &    4.641        & \textbf{0.044}           \\

\bottomrule
\end{tabular}
\end{adjustbox}

\caption{Quantitative comparison of ablated models. \textbf{Top:} Ablation on pipeline and training steps. (i) base model, (ii) finetuning on our data, (iii) conditioning on reference objects and background images, (iv) adding special grounding tokens, (v) adapting loss into final model. \textbf{Bottom:} Ablation on training data sources. (i) Professional In-the-Wild Images from Unsplash \cite{unsplash}, (ii) Professional Manual Designs, (iii) Pairs from Subject-200k \cite{chen2023subject}, (iv) Synthetic Side-by-side Compositions, (v) All sources combined.}
\label{tab:ablations}
\vspace{-1.3em}
\end{table}

\vspace{-0.2em}
\subsection{Emerging Capabilities}
\label{sec:apps}
\vspace{-0.1em}
While {\ourdm} is trained for text-guided multi-object compositing, it exhibits several emerging capabilities:







\vspace{1mm}\noindent\textbf{(i) Creative Composite Layouts.} As shown in \cref{fig:teaser} (bottom), {\ourdm} can automatically arrange objects in plausible scenes without explicit layout guidance, extrapolating unspecified attributes like lighting and sizing. This capability streamlines multi-object compositing and inspires novel layouts, enhancing its value in creative applications. 

\begin{figure}
    \centering
    \includegraphics[width=\linewidth]{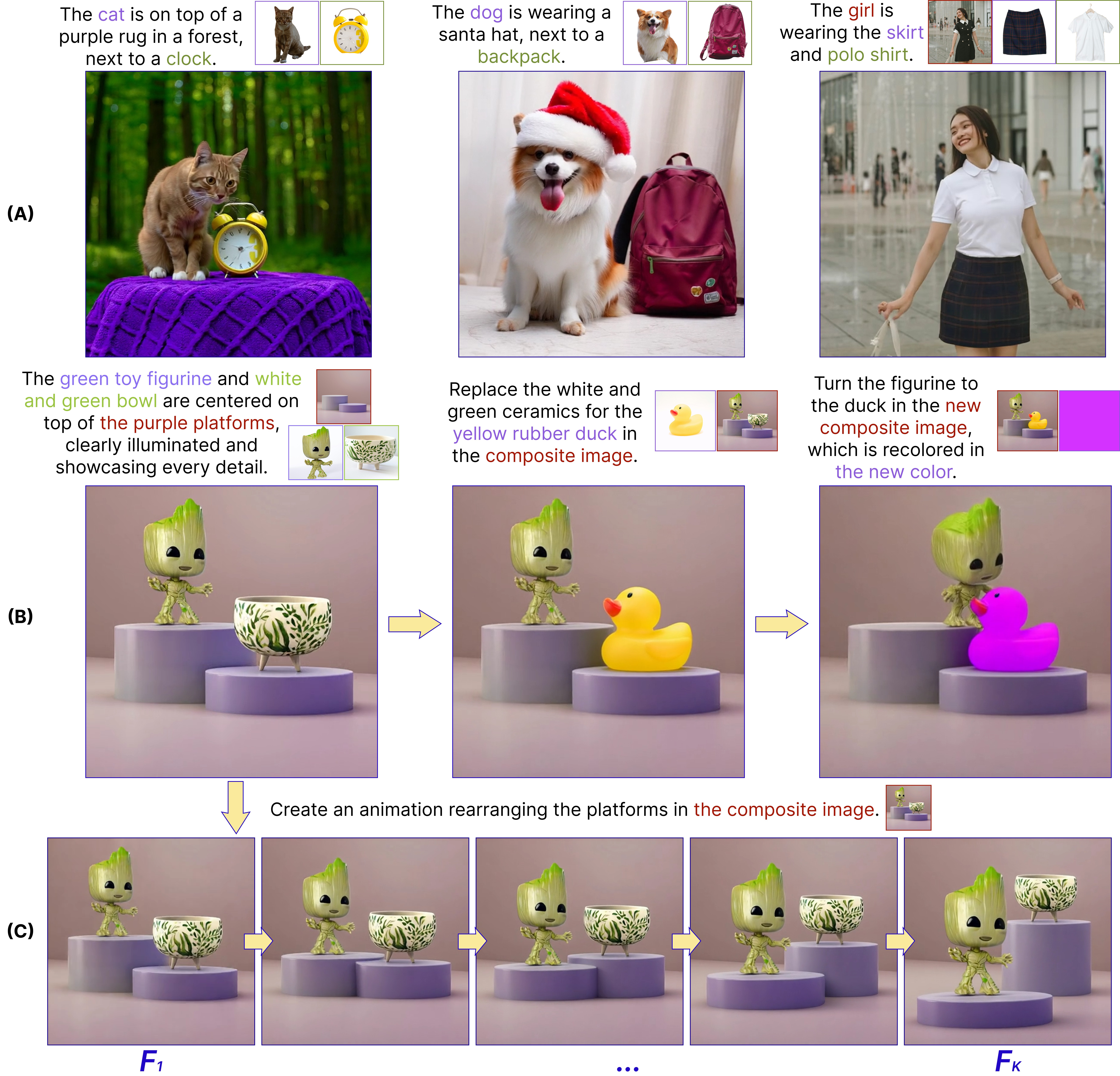}
    \caption{Emerging Capabilities. \textbf{A:} Text-guided generation with or without background, including interactive object addition (e.g., try-on). \textbf{B:} Image editing via text (replacement, rearrangement), object compositing, or color specification. \textbf{C:} Utilization of entire generated animation as output, beyond just the final frame.}
    \label{fig:apps}
    \vspace{-1.3em}
\end{figure}

\vspace{1mm}\noindent\textbf{(ii) Multi-Entity Subject-Driven Image Generation.} Our model can follow text input to create photorealistic scenes (\cref{fig:apps}A-left) even without providing a background image. It can also generate interacting elements (\cref{fig:apps}A-middle). 

\vspace{1mm}\noindent\textbf{(iii) Virtual Try-On.}  {\ourdm}'s ability to integrate new objects into existing scenes enables applications such as virtual clothing try-on, as demonstrated in \cref{fig:apps}A-right. 

\vspace{1mm}\noindent\textbf{(iv) Image Editing.} {\ourdm}'s versatility extends to image editing, leveraging text guidance, identity preservation, and fine-grained color control (\cref{fig:apps}B). Users can apply edits across these modalities, sequentially or simultaneously, enabling tasks from simple color adjustments to complex compositional changes while maintaining image integrity.

\vspace{1mm}\noindent\textbf{(v) Video Generation. } A key advantage of our video-based approach is the possibility of using the entire output sequence, instead of the final frame. As seen in \cref{fig:apps}C, {\ourdm} can produce short, consistent videos for edits or scene completions, usable as animated creative content. 

\vspace{-0.3em}
\subsection{Limitations}
\label{sec:limitations}
\vspace{-0.3em}
Although our method is optimized for preserving all distinct identities in the scene, achieving a natural and seamless composite image often requires reposing one or multiple objects. While minor reposing can be done without synthesizing a novel view, more significant repositioning necessitates completing hidden parts of the object. This completion, while plausible, might not be entirely consistent with the object in real life. To address this limitation in future work, the model could be augmented to accept multiple images of each object, providing more comprehensive information for accurate object reconstruction and reposing. 



\vspace{-0.3em}
\section{Conclusions}
\label{sec:conclusions}
\vspace{-0.3em}

In this paper, we introduce {\ourdm}, a novel approach for multi-object compositing that excels in identity preservation, background consistency and fine-grained color control. Leveraging video-based priors and a carefully curated dataset, our model combines multiple objects in a scene with text guidance, offering versatility in its applications. {\ourdm}'s capabilities extend to image refinement and video generation, providing an efficient solution for high-quality compositing without extensive manual input.


{
    \small
    \bibliographystyle{ieeenat_fullname}
    \bibliography{main}

@String(CVPR= {IEEE Conf. Comput. Vis. Pattern Recog.})

@String(ICCV= {Int. Conf. Comput. Vis.})

@String(ECCV= {Eur. Conf. Comput. Vis.})

@String(ICLR = {Int. Conf. Learn. Represent.})

@String(CVPR  = {CVPR})

@String(ICCV  = {ICCV})

@String(ECCV  = {ECCV})

@String(ICLR  = {ICLR})

@article{yuan2025opens2v,
  title={Opens2v-nexus: A detailed benchmark and million-scale dataset for subject-to-video generation},
  author={Yuan, Shenghai and He, Xianyi and Deng, Yufan and Ye, Yang and Huang, Jinfa and Lin, Bin and Luo, Jiebo and Yuan, Li},
  journal={arXiv preprint arXiv:2505.20292},
  year={2025}
}

@misc{unsplash,
    key = {https://unsplash.com/}
}

@inproceedings{liu2024grounding,
  title={Grounding dino: Marrying dino with grounded pre-training for open-set object detection},
  author={Liu, Shilong and Zeng, Zhaoyang and Ren, Tianhe and Li, Feng and Zhang, Hao and Yang, Jie and Jiang, Qing and Li, Chunyuan and Yang, Jianwei and Su, Hang and others},
  booktitle={ECCV},
  year={2024}
}

@inproceedings{kirillov2023segment,
  title={Segment anything},
  author={Kirillov, Alexander and Mintun, Eric and Ravi, Nikhila and Mao, Hanzi and Rolland, Chloe and Gustafson, Laura and Xiao, Tete and Whitehead, Spencer and Berg, Alexander C and Lo, Wan-Yen and others},
  booktitle={ICCV},
  year={2023}
}

@article{ren2024grounded,
  title={Grounded sam: Assembling open-world models for diverse visual tasks},
  author={Ren, Tianhe and Liu, Shilong and Zeng, Ailing and Lin, Jing and Li, Kunchang and Cao, He and Chen, Jiayu and Huang, Xinyu and Chen, Yukang and Yan, Feng and others},
  journal={arXiv preprint arXiv:2401.14159},
  year={2024}
}

@article{yu2023inpaint,
  title={Inpaint anything: Segment anything meets image inpainting},
  author={Yu, Tao and Feng, Runseng and Feng, Ruoyu and Liu, Jinming and Jin, Xin and Zeng, Wenjun and Chen, Zhibo},
  journal={arXiv preprint arXiv:2304.06790},
  year={2023}
}

@inproceedings{collins2022abo,
  title={Abo: Dataset and benchmarks for real-world 3d object understanding},
  author={Collins, Jasmine and Goel, Shubham and Deng, Kenan and Luthra, Achleshwar and Xu, Leon and Gundogdu, Erhan and Zhang, Xi and Vicente, Tomas F Yago and Dideriksen, Thomas and Arora, Himanshu and others},
  booktitle={CVPR},
  year={2022}
}

@inproceedings{tan2024ominicontrol,
  title={Ominicontrol: Minimal and universal control for diffusion transformer},
  author={Tan, Zhenxiong and Liu, Songhua and Yang, Xingyi and Xue, Qiaochu and Wang, Xinchao},
  booktitle={ICCV},
  year={2025}
}

@article{wang2025wan,
  title={Wan: Open and Advanced Large-Scale Video Generative Models},
  author={Wang, Ang and Ai, Baole and Wen, Bin and Mao, Chaojie and Xie, Chen-Wei and Chen, Di and Yu, Feiwu and Zhao, Haiming and Yang, Jianxiao and Zeng, Jianyuan and others},
  journal={CoRR},
  year={2025}
}

@article{hu2022lora,
  title={Lora: Low-rank adaptation of large language models.},
  author={Hu, Edward J and Shen, Yelong and Wallis, Phillip and Allen-Zhu, Zeyuan and Li, Yuanzhi and Wang, Shean and Wang, Lu and Chen, Weizhu and others},
  journal={ICLR},
  year={2022}
}

@inproceedings{yu2025objectmover,
  title={Objectmover: Generative object movement with video prior},
  author={Yu, Xin and Wang, Tianyu and Kim, Soo Ye and Guerrero, Paul and Chen, Xi and Liu, Qing and Lin, Zhe and Qi, Xiaojuan},
  booktitle={CVPR},
  year={2025}
}

@article{ju2025editverse,
  title={EditVerse: Unifying Image and Video Editing and Generation with In-Context Learning},
  author={Ju, Xuan and Wang, Tianyu and Zhou, Yuqian and Zhang, He and Liu, Qing and Zhao, Nanxuan and Zhang, Zhifei and Li, Yijun and Cai, Yuanhao and Liu, Shaoteng and others},
  journal={arXiv preprint arXiv:2509.20360},
  year={2025}
}

@inproceedings{radford2021learningCLIP,
  title={Learning transferable visual models from natural language supervision},
  author={Radford, Alec and Kim, Jong Wook and Hallacy, Chris and Ramesh, Aditya and Goh, Gabriel and Agarwal, Sandhini and Sastry, Girish and Askell, Amanda and Mishkin, Pamela and Clark, Jack and others},
  booktitle={ICML},
  year={2021}
}

@article{raffel2020exploringT5,
  title={Exploring the limits of transfer learning with a unified text-to-text transformer},
  author={Raffel, Colin and Shazeer, Noam and Roberts, Adam and Lee, Katherine and Narang, Sharan and Matena, Michael and Zhou, Yanqi and Li, Wei and Liu, Peter J},
  journal={JMLR},
  year={2020}
}

@inproceedings{mcqueen1967somekmeans,
  title={Some methods of classification and analysis of multivariate observations},
  author={McQueen, James B},
  booktitle={Proc. of 5th Berkeley Symposium on Math. Stat. and Prob.},
  year={1967}
}

@article{bai2025qwen2,
  title={Qwen2. 5-vl technical report},
  author={Bai, Shuai and Chen, Keqin and Liu, Xuejing and Wang, Jialin and Ge, Wenbin and Song, Sibo and Dang, Kai and Wang, Peng and Wang, Shijie and Tang, Jun and others},
  journal={arXiv preprint arXiv:2502.13923},
  year={2025}
}

@article{seedream2025seedream,
  title={Seedream 4.0: Toward next-generation multimodal image generation},
  author={Seedream, Team and Chen, Yunpeng and Gao, Yu and Gong, Lixue and Guo, Meng and Guo, Qiushan and Guo, Zhiyao and Hou, Xiaoxia and Huang, Weilin and Huang, Yixuan and others},
  journal={arXiv preprint arXiv:2509.20427},
  year={2025}
}

@article{labs2025flux,
  title={FLUX. 1 Kontext: Flow Matching for In-Context Image Generation and Editing in Latent Space},
  author={Labs, Black Forest and Batifol, Stephen and Blattmann, Andreas and Boesel, Frederic and Consul, Saksham and Diagne, Cyril and Dockhorn, Tim and English, Jack and English, Zion and Esser, Patrick and others},
  journal={arXiv preprint arXiv:2506.15742},
  year={2025}
}

@inproceedings{lipmanflow,
    title={Flow Matching for Generative Modeling},
    author={Yaron Lipman and Ricky T. Q. Chen and Heli Ben-Hamu and Maximilian Nickel and Matthew Le},
    booktitle={ICLR},
    year={2023}
}

@inproceedings{hessel2021clipscore,
  title={{CLIPScore:} A Reference-free Evaluation Metric for Image Captioning},
  author={Hessel, Jack and Holtzman, Ari and Forbes, Maxwell and Bras, Ronan Le and Choi, Yejin},
  booktitle={EMNLP},
  year={2021}
}

@article{oquab2024dinov2,
  title={DINOv2: Learning Robust Visual Features without Supervision},
  author={Oquab, Maxime and Darcet, Timoth{\'e}e and Moutakanni, Th{\'e}o and Vo, Huy and Szafraniec, Marc and Khalidov, Vasil and Fernandez, Pierre and Haziza, Daniel and Massa, Francisco and El-Nouby, Alaaeldin and others},
  journal={TMLR},
  year={2024}
}

@inproceedings{peng2024dreambench,
  author={Yuang Peng and Yuxin Cui and Haomiao Tang and Zekun Qi and Runpei Dong and Jing Bai and Chunrui Han and Zheng Ge and Xiangyu Zhang and Shu-Tao Xia},
  title={DreamBench++: A Human-Aligned Benchmark for Personalized Image Generation},
  booktitle={ICLR},
  year={2025},
}

@inproceedings{song2023objectstitch,
  title={Objectstitch: Object compositing with diffusion model},
  author={Song, Yizhi and Zhang, Zhifei and Lin, Zhe and Cohen, Scott and Price, Brian and Zhang, Jianming and Kim, Soo Ye and Aliaga, Daniel},
  booktitle={CVPR},
  year={2023}
}

@inproceedings{chen2024anydoor,
  title={Anydoor: Zero-shot object-level image customization},
  author={Chen, Xi and Huang, Lianghua and Liu, Yu and Shen, Yujun and Zhao, Deli and Zhao, Hengshuang},
  booktitle={CVPR},
  year={2024}
}

@inproceedings{tarres2025multitwine,
  title={Multitwine: Multi-object compositing with text and layout control},
  author={Tarr{\'e}s, Gemma Canet and Lin, Zhe and Zhang, Zhifei and Zhang, He and Gilbert, Andrew and Collomosse, John and Kim, Soo Ye},
  booktitle={CVPR},
  year={2025}
}

@inproceedings{canet2024thinking,
  title={Thinking outside the bbox: Unconstrained generative object compositing},
  author={Canet Tarr{\'e}s, Gemma and Lin, Zhe and Zhang, Zhifei and Zhang, Jianming and Song, Yizhi and Ruta, Dan and Gilbert, Andrew and Collomosse, John and Kim, Soo Ye},
  booktitle={ECCV},
  year={2024},
}

@inproceedings{yang2023paint,
  title={Paint by example: Exemplar-based image editing with diffusion models},
  author={Yang, Binxin and Gu, Shuyang and Zhang, Bo and Zhang, Ting and Chen, Xuejin and Sun, Xiaoyan and Chen, Dong and Wen, Fang},
  booktitle={CVPR},
  year={2023}
}

@article{wu2025uno,
  title={Less-to-more generalization: Unlocking more controllability by in-context generation},
  author={Wu, Shaojin and Huang, Mengqi and Wu, Wenxu and Cheng, Yufeng and Ding, Fei and He, Qian},
  journal={arXiv preprint arXiv:2504.02160},
  year={2025}
}

@inproceedings{cai2025dsd,
  title={Diffusion self-distillation for zero-shot customized image generation},
  author={Cai, Shengqu and Chan, Eric Ryan and Zhang, Yunzhi and Guibas, Leonidas and Wu, Jiajun and Wetzstein, Gordon},
  booktitle={CVPR},
  year={2025}
}

@inproceedings{xiao2025omnigen,
  title={Omnigen: Unified image generation},
  author={Xiao, Shitao and Wang, Yueze and Zhou, Junjie and Yuan, Huaying and Xing, Xingrun and Yan, Ruiran and Li, Chaofan and Wang, Shuting and Huang, Tiejun and Liu, Zheng},
  booktitle={CVPR},
  year={2025}
}

@article{wang2024ms,
  title={Ms-diffusion: Multi-subject zero-shot image personalization with layout guidance},
  author={Wang, Xierui and Fu, Siming and Huang, Qihan and He, Wanggui and Jiang, Hao},
  journal={arXiv preprint arXiv:2406.07209},
  year={2024}
}

@article{jiang2025vace,
  title={Vace: All-in-one video creation and editing},
  author={Jiang, Zeyinzi and Han, Zhen and Mao, Chaojie and Zhang, Jingfeng and Pan, Yulin and Liu, Yu},
  journal={arXiv preprint arXiv:2503.07598},
  year={2025}
}

@article{wu2025qwen,
  title={Qwen-image technical report},
  author={Wu, Chenfei and Li, Jiahao and Zhou, Jingren and Lin, Junyang and Gao, Kaiyuan and Yan, Kun and Yin, Sheng-ming and Bai, Shuai and Xu, Xiao and Chen, Yilei and others},
  journal={arXiv preprint arXiv:2508.02324},
  year={2025}
}

@article{comanici2025gemini,
  title={Gemini 2.5: Pushing the frontier with advanced reasoning, multimodality, long context, and next generation agentic capabilities},
  author={Comanici, Gheorghe and Bieber, Eric and Schaekermann, Mike and Pasupat, Ice and Sachdeva, Noveen and Dhillon, Inderjit and Blistein, Marcel and Ram, Ori and Zhang, Dan and Rosen, Evan and others},
  journal={arXiv preprint arXiv:2507.06261},
  year={2025}
}

@inproceedings{guerreiro2023pct,
  title={Pct-net: Full resolution image harmonization using pixel-wise color transformations},
  author={Guerreiro, Julian Jorge Andrade and Nakazawa, Mitsuru and Stenger, Bj{\"o}rn},
  booktitle={CVPR},
  year={2023}
}

@inproceedings{jiang2021ssh,
  title={Ssh: A self-supervised framework for image harmonization},
  author={Jiang, Yifan and Zhang, He and Zhang, Jianming and Wang, Yilin and Lin, Zhe and Sunkavalli, Kalyan and Chen, Simon and Amirghodsi, Sohrab and Kong, Sarah and Wang, Zhangyang},
  booktitle={ICCV},
  year={2021}
}

@incollection{perez2023poisson,
  title={Poisson image editing},
  author={P{\'e}rez, Patrick and Gangnet, Michel and Blake, Andrew},
  booktitle={Seminal Graphics Papers: Pushing the Boundaries, Volume 2},
  pages={577--582},
  year={2023}
}

@inproceedings{zhang2021deep,
  title={Deep image compositing},
  author={Zhang, He and Zhang, Jianming and Perazzi, Federico and Lin, Zhe and Patel, Vishal M},
  booktitle={WACV},
  year={2021}
}

@inproceedings{song2024imprint,
  title={Imprint: Generative object compositing by learning identity-preserving representation},
  author={Song, Yizhi and Zhang, Zhifei and Lin, Zhe and Cohen, Scott and Price, Brian and Zhang, Jianming and Kim, Soo Ye and Zhang, He and Xiong, Wei and Aliaga, Daniel},
  booktitle={CVPR},
  year={2024}
}

@article{chen2023subject,
  title={Subject-driven text-to-image generation via apprenticeship learning},
  author={Chen, Wenhu and Hu, Hexiang and Li, Yandong and Ruiz, Nataniel and Jia, Xuhui and Chang, Ming-Wei and Cohen, William W},
  journal={NeurIPS},
  year={2023}
}

@inproceedings{ruiz2023dreambooth,
  title={Dreambooth: Fine tuning text-to-image diffusion models for subject-driven generation},
  author={Ruiz, Nataniel and Li, Yuanzhen and Jampani, Varun and Pritch, Yael and Rubinstein, Michael and Aberman, Kfir},
  booktitle={CVPR},
  year={2023}
}

@inproceedings{shi2024instantbooth,
  title={Instantbooth: Personalized text-to-image generation without test-time finetuning},
  author={Shi, Jing and Xiong, Wei and Lin, Zhe and Jung, Hyun Joon},
  booktitle={CVPR},
  year={2024}
}

@article{gal2022image,
  title={An image is worth one word: Personalizing text-to-image generation using textual inversion},
  author={Gal, Rinon and Alaluf, Yuval and Atzmon, Yuval and Patashnik, Or and Bermano, Amit H and Chechik, Gal and Cohen-Or, Daniel},
  journal={arXiv preprint arXiv:2208.01618},
  year={2022}
}

@inproceedings{kumari2023multi,
  title={Multi-concept customization of text-to-image diffusion},
  author={Kumari, Nupur and Zhang, Bingliang and Zhang, Richard and Shechtman, Eli and Zhu, Jun-Yan},
  booktitle={CVPR},
  year={2023}
}

@article{li2023blip,
  title={Blip-diffusion: Pre-trained subject representation for controllable text-to-image generation and editing},
  author={Li, Dongxu and Li, Junnan and Hoi, Steven},
  journal={NeurIPS},
  year={2023}
}

@article{ye2023ip,
  title={Ip-adapter: Text compatible image prompt adapter for text-to-image diffusion models},
  author={Ye, Hu and Zhang, Jun and Liu, Sibo and Han, Xiao and Yang, Wei},
  journal={arXiv preprint arXiv:2308.06721},
  year={2023}
}

@inproceedings{brooks2023instructpix2pix,
  title={Instructpix2pix: Learning to follow image editing instructions},
  author={Brooks, Tim and Holynski, Aleksander and Efros, Alexei A},
  booktitle={CVPR},
  year={2023}
}

@article{zhang2023magicbrush,
  title={Magicbrush: A manually annotated dataset for instruction-guided image editing},
  author={Zhang, Kai and Mo, Lingbo and Chen, Wenhu and Sun, Huan and Su, Yu},
  journal={NeurIPS},
  year={2023}
}

@article{hertz2022prompt,
  title={Prompt-to-prompt image editing with cross attention control},
  author={Hertz, Amir and Mokady, Ron and Tenenbaum, Jay and Aberman, Kfir and Pritch, Yael and Cohen-Or, Daniel},
  journal={arXiv preprint arXiv:2208.01626},
  year={2022}
}

@inproceedings{nanopenvid,
  title={OpenVid-1M: A Large-Scale High-Quality Dataset for Text-to-video Generation},
  author={Nan, Kepan and Xie, Rui and Zhou, Penghao and Fan, Tiehan and Yang, Zhenheng and Chen, Zhijie and Li, Xiang and Yang, Jian and Tai, Ying},
  booktitle={ICLR},
  year={2025}
}

@inproceedings{wanginternvid,
  title={InternVid: A Large-scale Video-Text Dataset for Multimodal Understanding and Generation},
  author={Wang, Yi and He, Yinan and Li, Yizhuo and Li, Kunchang and Yu, Jiashuo and Ma, Xin and Li, Xinhao and Chen, Guo and Chen, Xinyuan and Wang, Yaohui and others},
  booktitle={ICLR},
  year={2025}
}

@inproceedings{chen2024panda,
  title={Panda-70m: Captioning 70m videos with multiple cross-modality teachers},
  author={Chen, Tsai-Shien and Siarohin, Aliaksandr and Menapace, Willi and Deyneka, Ekaterina and Chao, Hsiang-wei and Jeon, Byung Eun and Fang, Yuwei and Lee, Hsin-Ying and Ren, Jian and Yang, Ming-Hsuan and others},
  booktitle={CVPR},
  year={2024}
}

@article{zi2025senorita,
  title={Se$\backslash$\~{} norita-2M: A High-Quality Instruction-based Dataset for General Video Editing by Video Specialists},
  author={Zi, Bojia and Ruan, Penghui and Chen, Marco and Qi, Xianbiao and Hao, Shaozhe and Zhao, Shihao and Huang, Youze and Liang, Bin and Xiao, Rong and Wong, Kam-Fai},
  journal={arXiv preprint arXiv:2502.06734},
  year={2025}
}

@article{hu2024vivid,
  title={VIVID-10M: A Dataset and Baseline for Versatile and Interactive Video Local Editing},
  author={Hu, Jiahao and Zhong, Tianxiong and Wang, Xuebo and Jiang, Boyuan and Tian, Xingye and Yang, Fei and Wan, Pengfei and Zhang, Di},
  journal={arXiv preprint arXiv:2411.15260},
  year={2024}
}

@inproceedings{qin2024instructvid2vid,
  title={Instructvid2vid: Controllable video editing with natural language instructions},
  author={Qin, Bosheng and Li, Juncheng and Tang, Siliang and Chua, Tat-Seng and Zhuang, Yueting},
  booktitle={ICMR},
  year={2024},
}

@inproceedings{lin2025realgeneral,
  title={Realgeneral: Unifying visual generation via temporal in-context learning with video models},
  author={Lin, Yijing and Huang, Mengqi and Zhuang, Shuhan and Mao, Zhendong},
  booktitle={ICCV},
  year={2025}
}

@article{wu2025chronoedit,
  title={ChronoEdit: Towards Temporal Reasoning for Image Editing and World Simulation},
  author={Wu, Jay Zhangjie and Ren, Xuanchi and Shen, Tianchang and Cao, Tianshi and He, Kai and Lu, Yifan and Gao, Ruiyuan and Xie, Enze and Lan, Shiyi and Alvarez, Jose M and others},
  journal={arXiv preprint arXiv:2510.04290},
  year={2025}
}

@misc{openai2025gptoss120bgptoss20bmodel,
      title={gpt-oss-120b \& gpt-oss-20b Model Card}, 
      author={OpenAI},
      year={2025},
      eprint={2508.10925},
      archivePrefix={arXiv},
      primaryClass={cs.CL},
      url={https://arxiv.org/abs/2508.10925}, 
}
}

\clearpage
\appendix
\renewcommand{\thetable}{S.\arabic{table}}  
\renewcommand{\thefigure}{S.\arabic{figure}} 
\renewcommand{\theequation}{S.\arabic{equation}}
\renewcommand{\thesection}{S.\arabic{section}}
\refstepcounter{figure} 
\refstepcounter{table} 
\refstepcounter{equation}
\setcounter{figure}{0}  
\setcounter{table}{0}  
\setcounter{equation}{0}
\definecolor{obj1}{rgb}{0.55, 0.35, 0.866}
\definecolor{obj2}{rgb}{0.46, 0.56, 0.22}
\definecolor{obj3}{rgb}{0.95, 0.5647, 0.0471}
\definecolor{bg}{rgb}{0.6471, 0.0863, 0.0118}

\clearpage
\setcounter{page}{1}
\maketitlesupplementary

This Supplementary Material provides complementary information and visuals on different aspects presented in the main paper. It includes additional visualizations of emerging capabilities (\cref{sec:sup-emerging}) and ablation studies (\cref{sec:sup-ablation}). Further experiments on video length and first frame initialization are presented in \cref{sec:sup-experiments}. Expanded information on training data and the evaluation set are detailed in \cref{sec:sup-training} and \cref{sec:sup-testset}, respectively. Comparisons with state of the art works are shown in \cref{sec:sup-soa}, while model limitations are discussed in \cref{sec:sup-limitations}.




\section{Emerging Capabilities}
\label{sec:sup-emerging}

\begin{figure*}
    \centering
    \includegraphics[width=\linewidth]{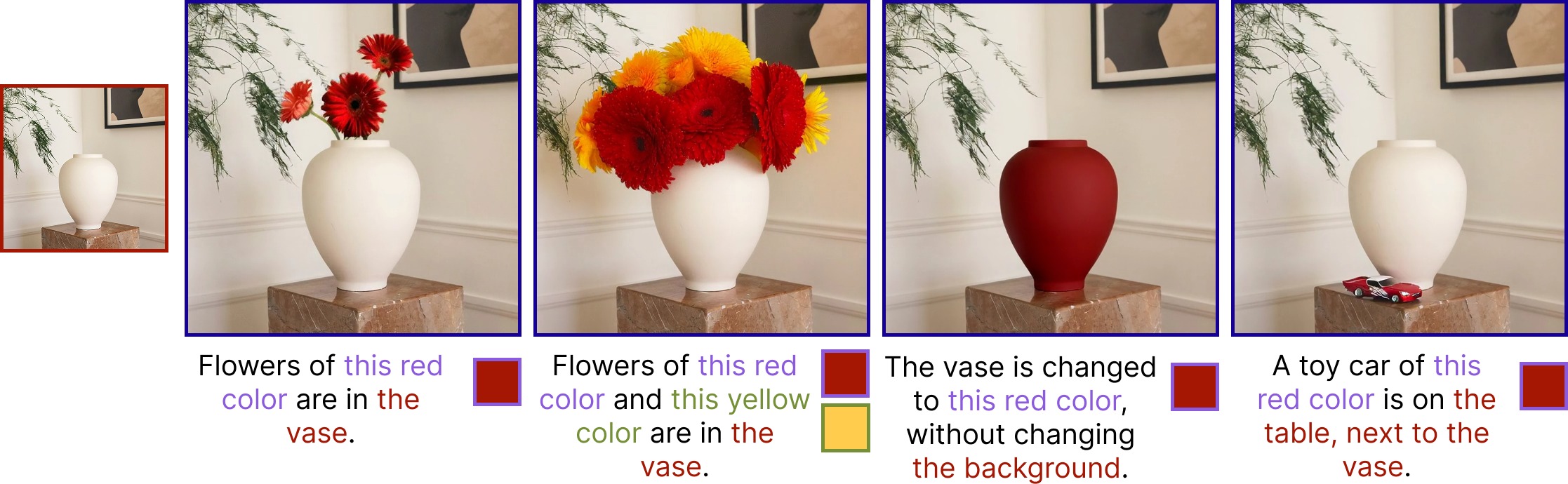}
    \caption{Text- and Color-Guided Image Editing. Four edits are applied to a single background image using text and color instructions.}
    \label{fig:sup-edit}
\end{figure*}
\begin{figure*}
    \centering
    \includegraphics[width=\linewidth]{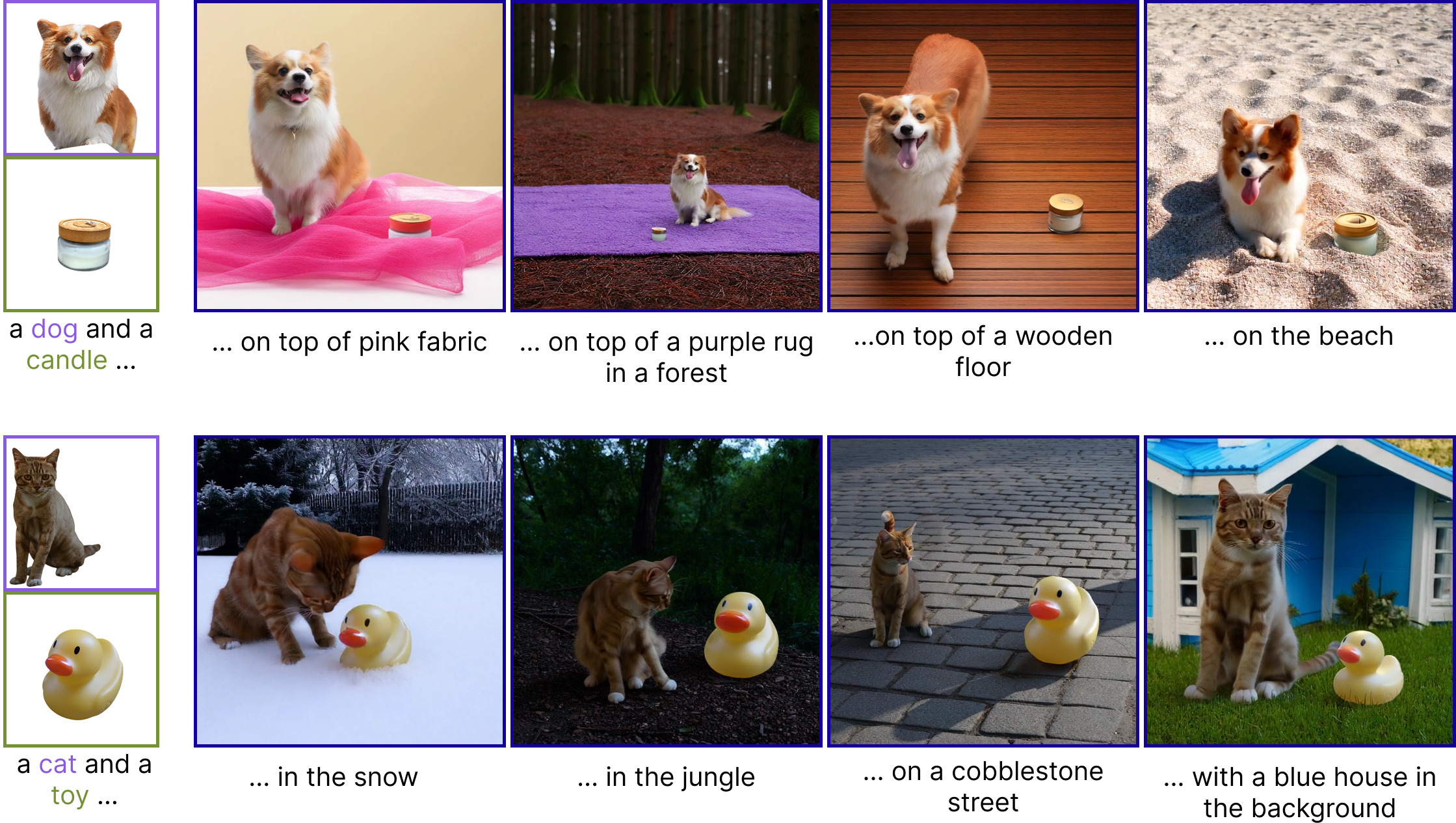}
    \caption{Subject-Guided Image Generation. {\ourdm} can generate images where objects are contextualized based on textual prompts.}
    \label{fig:sup-customization}
\end{figure*}
\begin{figure*}
    \centering
    \includegraphics[width=\linewidth]{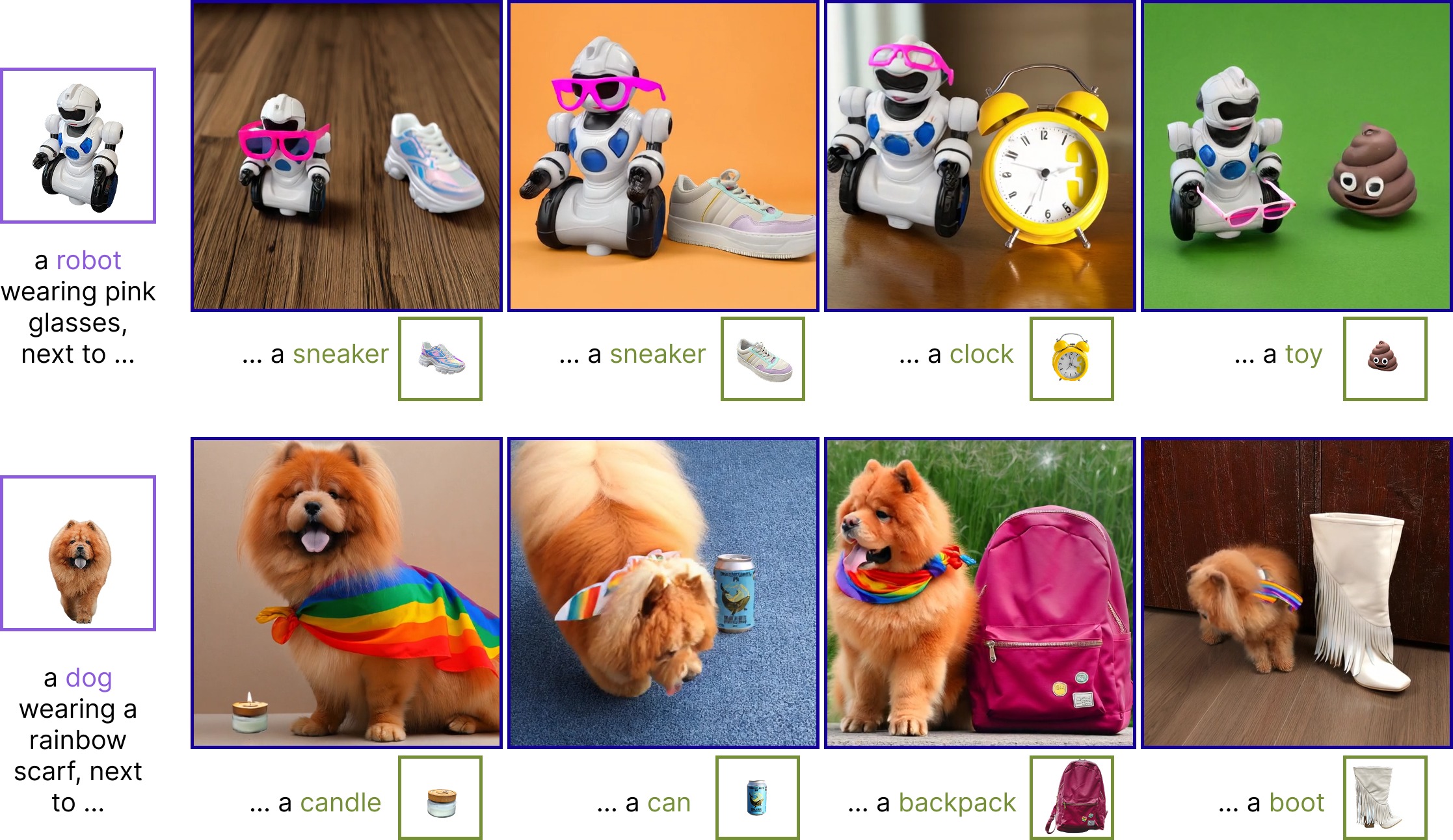}
    \caption{Subject-Guided Generation with Text-Guided Interacting Objects. {\ourdm} can generate background scenes and additional interacting objects (\eg, glasses, scarves) based on textual prompts, incorporating them with the main subjects.}
    \label{fig:sup-textinteract}
\end{figure*}

As explained in \refmainsec{4.3},
our model's capabilities extend beyond object compositing. By leveraging its ability to maintain background integrity, align with text instructions, and faithfully reproduce colors, it can perform text-based and color-based image editing, as demonstrated in \cref{fig:sup-edit}. Furthermore, when no background image is provided, the model can synthesize text-aligned backgrounds, as shown in \cref{fig:sup-customization}. It can even generate additional objects or props that interact with the main subjects, as illustrated in \cref{fig:sup-textinteract}. 


\section{Ablation Studies}
\label{sec:sup-ablation}

\begin{figure*}
    \centering
    \includegraphics[width=\linewidth]{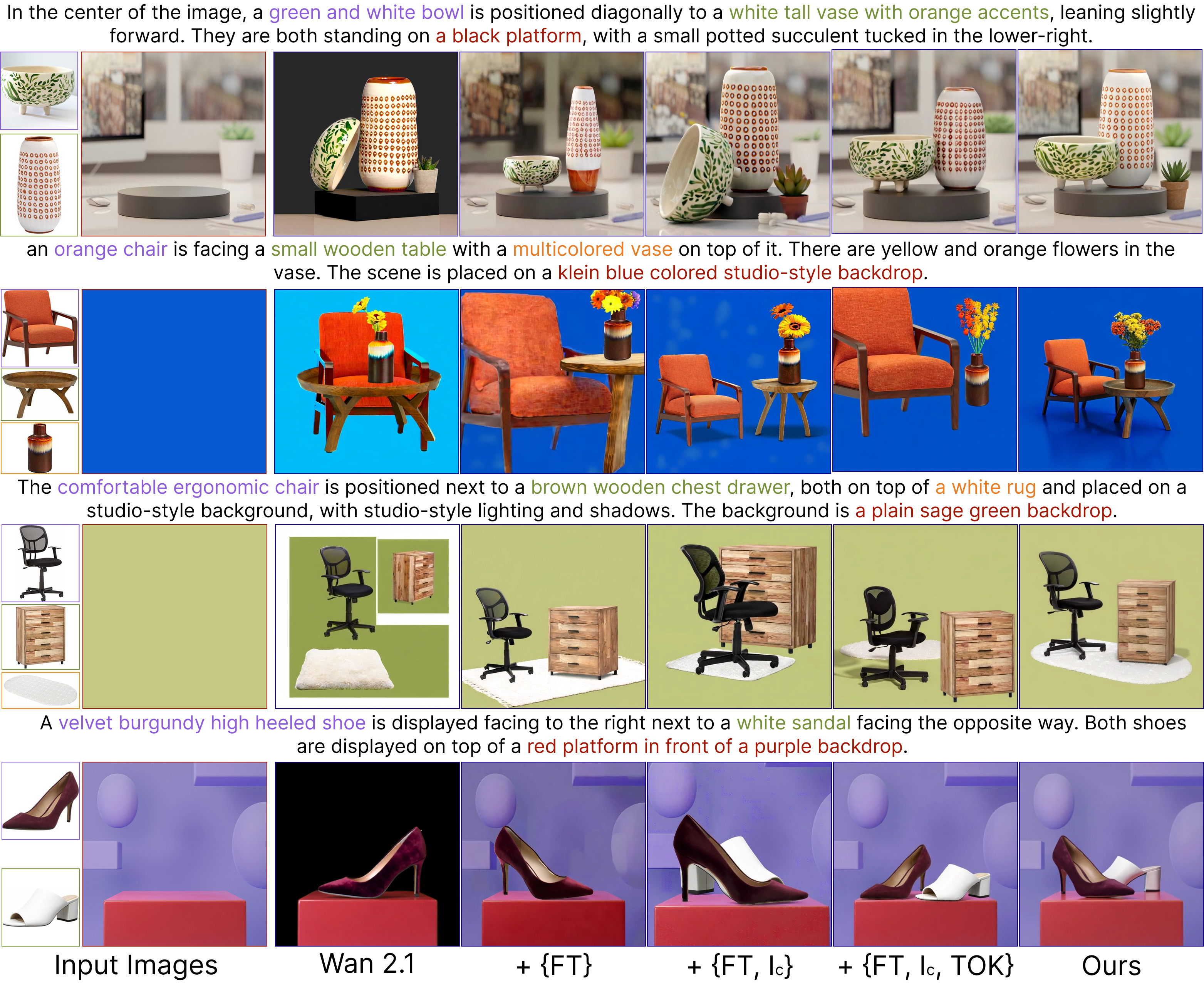}
    \caption{Training Strategy Ablation Studies. We visually compare generation on (i) base model \cite{wang2025wan}, (ii) finetuning on our data, (iii) conditioning on object images and background scene, (iv) adding special grounding tokens, and (v) adapting loss into our final model.}
    \label{fig:sup-ablationtraining}
\end{figure*}

\begin{figure*}
    \centering
    \includegraphics[width=\linewidth]{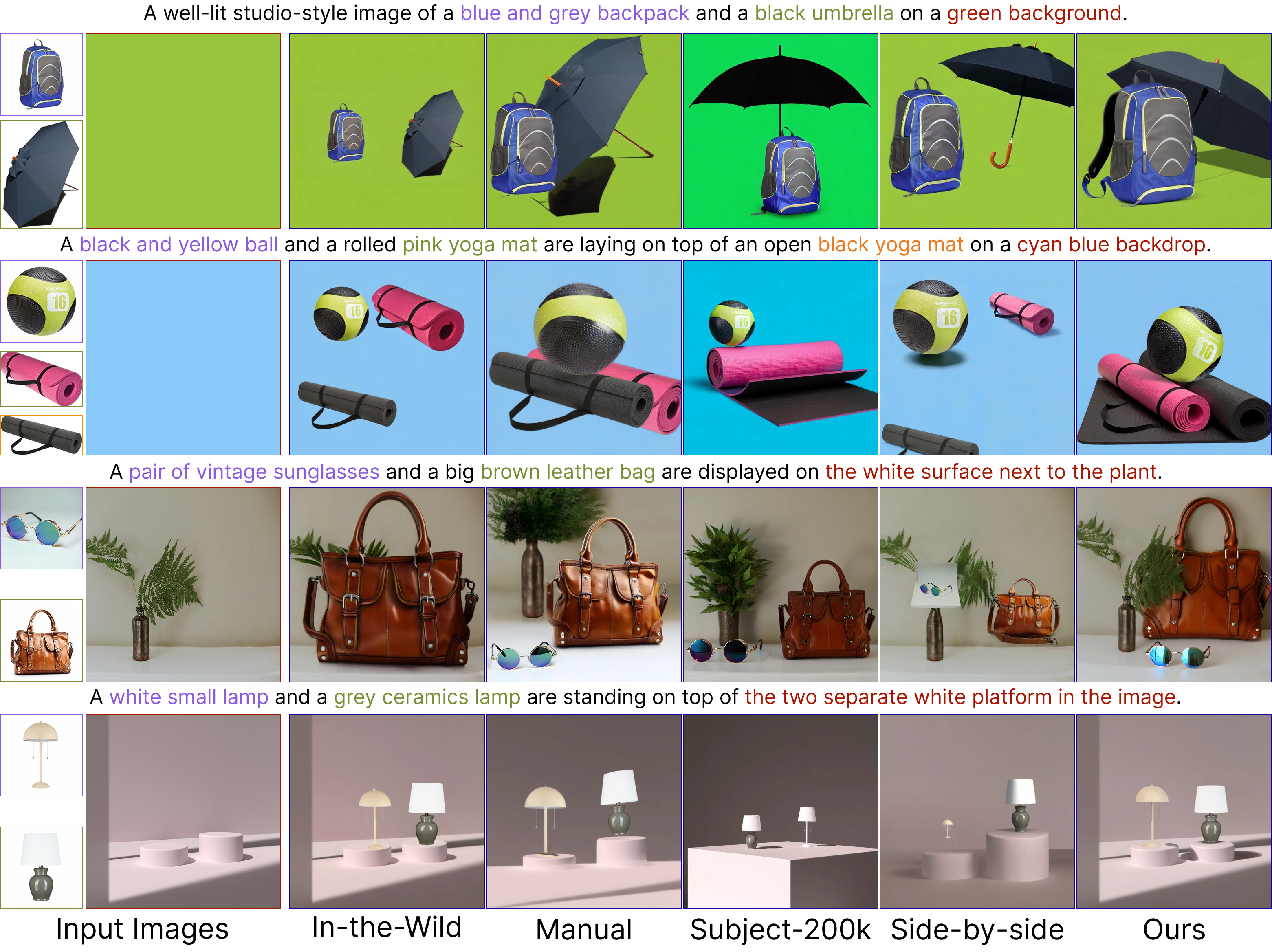}
    \caption{Training Data Ablation Studies. We visually compare training with (i) In-the-Wild Professional Multi-Object Images from Unsplash \cite{unsplash}, (ii) Professional Manual Designs, (iii) Pairs from Subject-200k \cite{chen2023subject}, (iv) Synthetic Side-by-side Compositions, and (v) Our model, simultaneously trained with all data sources.}
    \label{fig:sup-ablationdata}
\end{figure*}

\paragraph{(i) Training Strategy.} We visualize the effects of each additive change in our training pipeline in \cref{fig:sup-ablationtraining}. The base model struggles with our task due to the significant difference between evaluation and training data. Finetuning on our entire training set improves results but still yields blurry, low-quality images with poor identity preservation. Conditioning on object and background images enhances object identity preservation and consistency, though text instruction adherence remains challenging (\cref{fig:sup-ablationtraining} rows 1,3,4). Adding special text tokens for better text-visual input correlation improves text understanding and composite coherence. However, using noisy intermediate frames from Subject-200k \cite{chen2023subject} data leads to missing (\cref{fig:sup-ablationtraining} row 2) and distorted objects (\cref{fig:sup-ablationtraining} row 3). By using only the final frame from Subject-200k for training supervision in our final model, we reduce this noise, resulting in improved text-image alignment, more natural and complete scenes, and better fine-grained preservation of objects and backgrounds. 

\paragraph{(ii) Training Data.} \cref{fig:sup-ablationdata} compares models trained on different data sources. In-the-Wild images from Unsplash \cite{unsplash} offer good photorealistic background preservation but tend towards copy-pasting, especially on plain backgrounds, and struggle with textual instructions and reposing. Manually designed compositions provide appealing layouts on plain backgrounds but struggle with photorealistic backgrounds, object interactions, and text-image alignment. Subject-200k data greatly improves text alignment (\cref{fig:sup-ablationdata}, row 2) but performs poorly in background preservation and retaining specific object details. Side-by-side compositions provide a stronger bias towards object relighting and relative scaling, but struggle with drastic reposing and integrating objects in photorealistic scenes (\cref{fig:sup-ablationdata}, rows 3,4). Our final model effectively combines these complementary data sources, offering natural-looking scenes with coherently transformed and relighted objects, good text alignment, preserved object identities, and faithful background scenes, including fine-grained colors.

\section{Additional Experiments}
\label{sec:sup-experiments}

\paragraph{(i) Study on Number of Frames.} We train our model on 9-frame videos, balancing the need for sufficient length to learn object animation along synthetic trajectories with efficient use of training resources. At inference time, the model can generate videos of varying lengths, with the last frame always serving as the target image. \cref{tab:sup-numberframes} illustrates how different quantitative metrics change with video length, from 9 to 81 frames (the default in our base model \cite{wang2025wan}). Longer videos allow for greater transformations, improving text-image alignment but potentially compromising identity, background, and color fidelity. Conversely, shorter videos (9 or 17 frames) may not fully transition objects from initial random locations to the desired composition, resulting in lower text-image alignment. For our compositing experiments in the main paper and SupMat, we use 33 frames to achieve an optimal balance between identity preservation and text-image alignment. Notably, users can choose to adjust this trade-off by simply generating shorter or longer videos, providing flexibility in the model's application.

\begin{table}[t!]

\centering
\begin{adjustbox}{width=\linewidth}

\begin{tabular}{lcccccc}
\toprule
\textbf{\# Frames}                            & \textbf{CLIP-I$\uparrow$} & \textbf{DINO$\uparrow$} & \textbf{CLIP-T$\uparrow$} &  \textbf{MSE-BG$\downarrow$} & \textbf{Chamfer $\downarrow$} & \textbf{Missing  $\downarrow$} \\

\midrule 
9     &  0.694  &   \textbf{0.450}       &  0.330                                                                                             &  0.027   &  3.228        &   0.045    \\
 \midrule
17     &  0.688 &  0.444        &      0.330                                                                                         &  0.027   &   5.590       &  0.048     \\
 \midrule

33     &  \textbf{0.705} & 0.440          &            0.336                                                                                        &   \textbf{0.019}   &    4.641        &    \textbf{0.044}       \\
\midrule 
61     &  0.699 &   0.436       &    0.334                                                                                           &  0.032   &  \textbf{3.061}        &  0.054     \\
\midrule 
81     & 0.693  &     0.422     &    \textbf{0.337}                                                                                           &  0.043   &    5.500      &  0.057     \\

\bottomrule
\end{tabular}
\end{adjustbox}

\caption{Impact of Video Length on Image Quality Metrics. This study examines the effect of varying the number of generated video frames on the quality of the final output image. We quantitatively evaluate: identity preservation (CLIP-I, DINO), text alignment (CLIP-T), background preservation (MSE-BG), color fidelity (Chamfer), and object omission rate (Missing).} 
\label{tab:sup-numberframes}
\end{table}

\begin{figure}
    \centering
    \includegraphics[width=\linewidth]{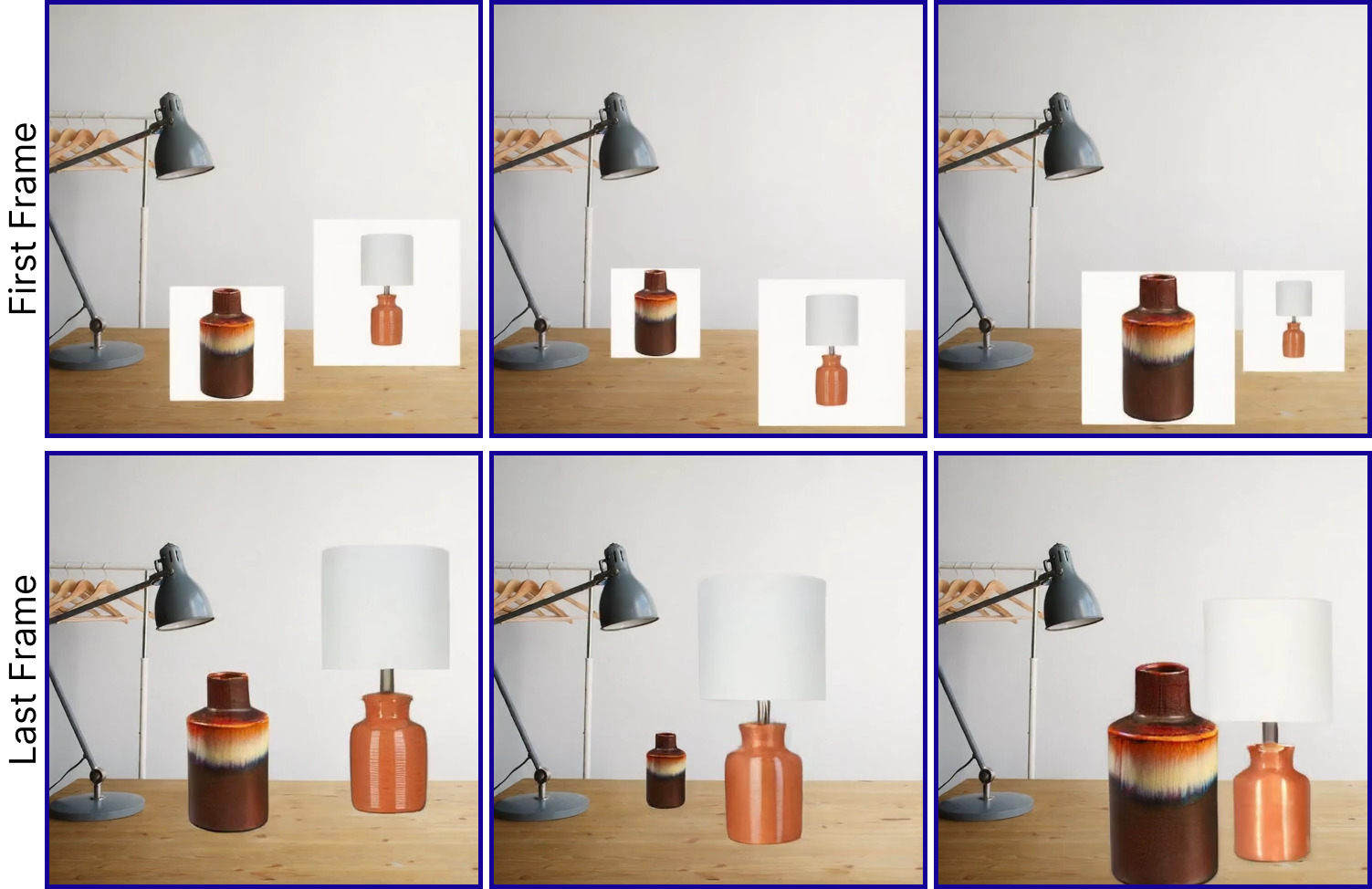}
    \caption{Effect of First Frame Initialization on Final Composition. When generating short videos, the initial object placement in the first frame can be used as rough guidance for the layout in the last frame, when combined with a complementary caption. \textbf{(Top):} First frame initialization; \textbf{(Bottom):} Last frame of generated 9-frame video, guided by ``The vase and the lamp are on the table".} 
    \label{fig:sup-firstframe}
\end{figure}

\paragraph{(ii) Study on First Frame Initialization.} As detailed in \cref{sec:sup-experiments} (i), our model's use of fewer frames at inference time limits the extent to which objects can move from their initial to final positions. While this might seem restrictive, it actually provides an additional layer of control. Users can take advantage of this characteristic by manually choosing how to arrange objects on the background to create the first frame, rather than relying solely on text descriptions to specify the desired layout. Due to the short video length, objects tend to remain close to their initial positions and scale, as shown in \cref{fig:sup-firstframe}. This method is only effective when the text caption aligns with the initial object arrangement or when no caption is provided. However, it's important to note that if the caption specifies a different layout, these textual instructions will override the first frame initialization. This feature offers users a visual means of controlling object placement while still preserving the model's ability to respond to text prompts. 

\section{Training Data Generation}
\label{sec:sup-training}

{\ourdm} is trained on video data which depicts objects following synthetic trajectories from initial random positions to form visually appealing, coherent layouts in the final frame. As detailed in \refmainsec{3.3},
these videos are generated from three distinct image sources: (i) Professional Multi-Object Images, (ii) a Subject-Driven Generation Paired Dataset, and (iii) Synthetic Side-by-side Compositions. This section provides additional information and visualizations for each of these sources, along with more details on the augmentations applied to the training data.


\subsection{Professional Multi-Object Images}

\begin{figure}
    \centering
    \includegraphics[width=\linewidth]{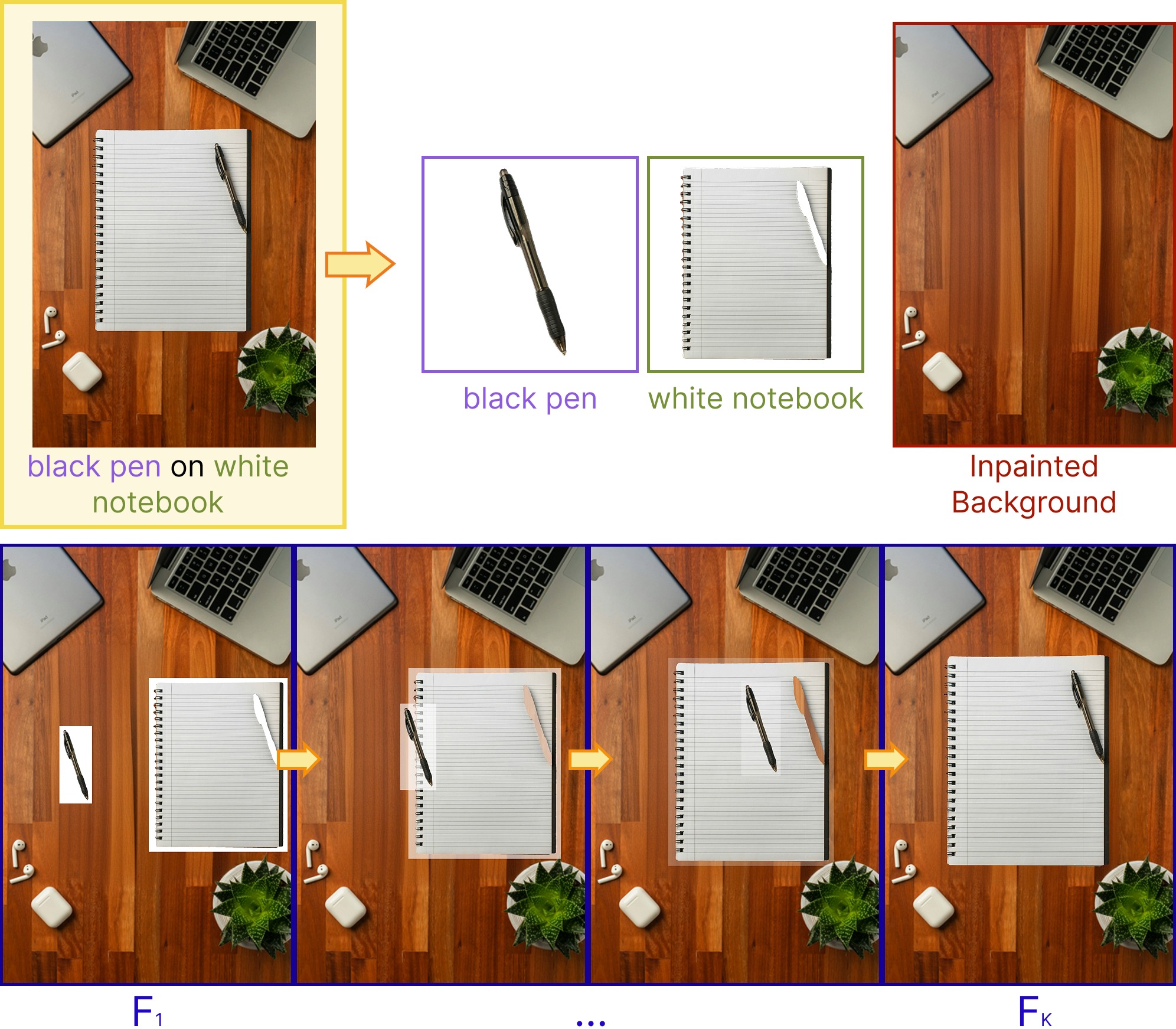}
    \caption{Training Data Generation Pipeline from In-the-Wild Images. Given an image-caption pair from Unsplash \cite{unsplash}, the process extracts relevant objects and an inpainted background. These elements are used to create the $F_{1..K}$ video frames for training.}
    \label{fig:sup-unsplash}
\end{figure}

\textbf{(a) In-the-Wild Images:} All 13,103 images labeled `Product Photography', or `Flat Lay'
on Unsplash \cite{unsplash} are parsed together with their captions (\eg, yellow box contents in \cref{fig:sup-unsplash}). For each image we first identify the relevant foreground objects by using the grounding pipeline of Grounding‑DINO \cite{liu2024grounding} together with SAM \cite{kirillov2023segment}, as in \cite{ren2024grounded}. The raw detections are then cleaned by (i) merging those with the same label, (ii) discarding duplicates with different labels, (iii) eliminating overlap by separating small objects largely interacting with larger ones (\eg, pen and notebook in \cref{fig:sup-unsplash}), and (iv) keeping only objects with mask‑coverage between 0.5\% and 80\% of image area. Images that end up with no valid detections are dropped, and the remaining backgrounds are restored with the inpainting method of \cite{yu2023inpaint} on the union of all foreground objects, using a dilation of $50$. Each retained object is extracted, randomly reordered, rescaled, and optionally warped with a mild perspective transform; it is then pasted onto a white box to emulate the unsegmented object image $I_i$. To build $F_1$, the first frame of the training video, we scatter these boxed objects across an empty canvas. The canvas background can be (i) plain white, (ii) a randomly chosen background, or (iii) the inpainted original background. Finally, a short video is rendered as in \cref{fig:sup-unsplash} (bottom): every object travels in a straight line at constant speed from its random start location to its target position in the professional composition, while simultaneously the white boxes fade out and, if a plain‑white background was used initially, the true background fades in. The last frame of the video, $F_K$, is thus the desired multi-object composite image, and the intermediate frames provide a temporally and spatially coherent progression used for training the model.

\begin{figure}
    \centering
    \includegraphics[width=\linewidth]{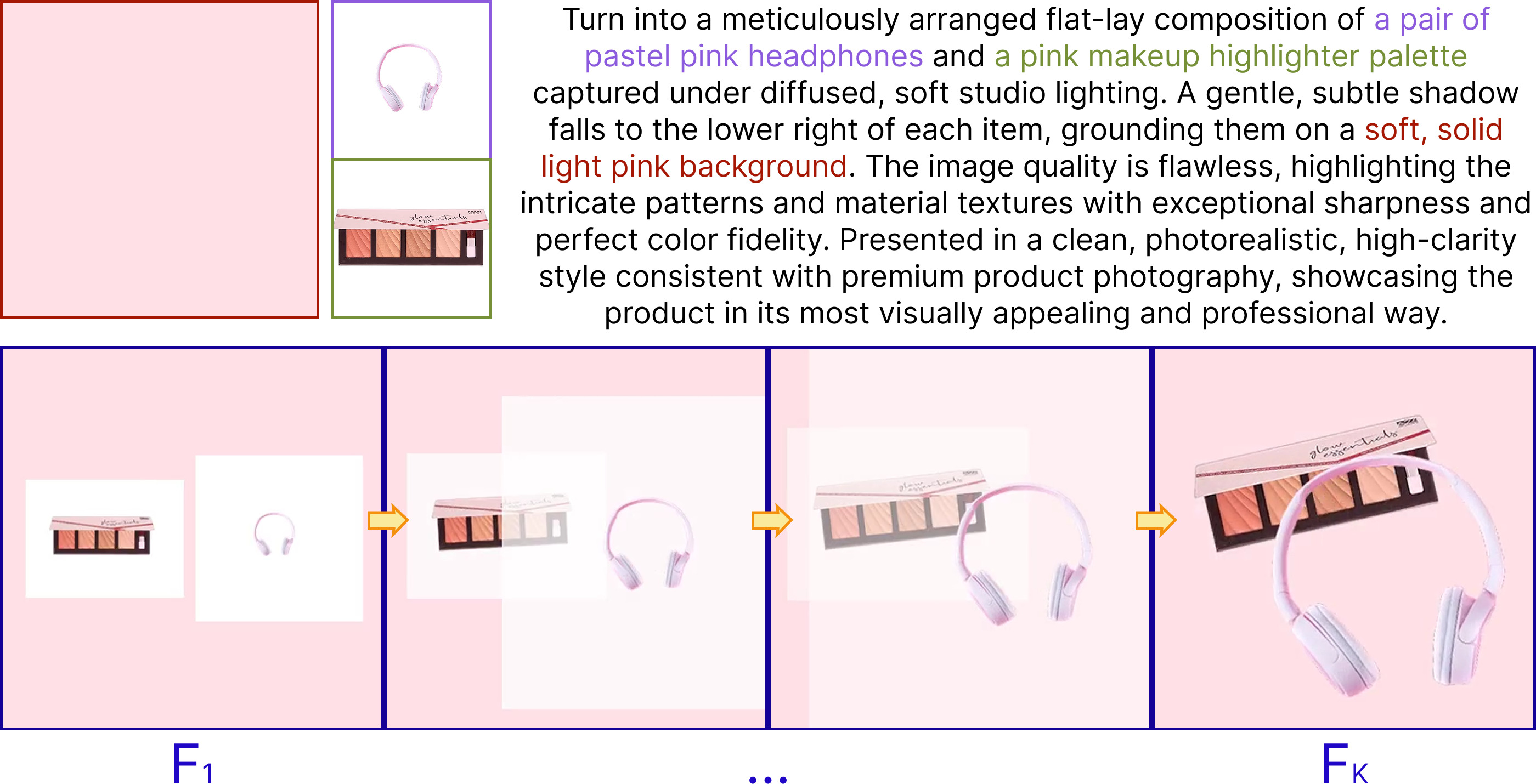}
    \caption{Training Data Generation Pipeline from Manual Designs. This process transforms a random arrangement of objects on a background image $F_1$ into a professionally designed layout $F_K$ over a short video sequence. Objects follow synthetic trajectories, while a template-based caption describes the animation.} 
    \label{fig:sup-figma}
\end{figure}

\textbf{(b) Manual Designs: } In-the-wild multi-object images often contain severe occlusions or cut‑off objects, making it impossible to recover their full true appearance (\eg, notebook in \cref{fig:sup-unsplash}). Since faithful detail preservation is essential in our task, we request professional designers to curate a small set of $\sim 400$ manually designed compositions. For each composition, we have access to a separate image of each object, showing its entire appearance. As visualized in \cref{fig:sup-figma}, given a background image and a set of high-resolution images for 2-5 objects, the designers provide a visually appealing image $F_K$, including all objects on the provided background. We synthesize the short video $F_{1..K}$ by randomly initializing the first frame, placing all unsegmented objects on the background image, and progressively rotate, translate and transform the objects to their final arrangement in $F_K$. The caption paired with each video is created by following one of a few pre-computed template prompts that describe the desired transformation and scene, completed with a short description of each individual object and background, individually generated using \cite{bai2025qwen2}.


\subsection{Subject-Driven Generation Paired Dataset}

\begin{figure}
    \centering
    \includegraphics[width=\linewidth]{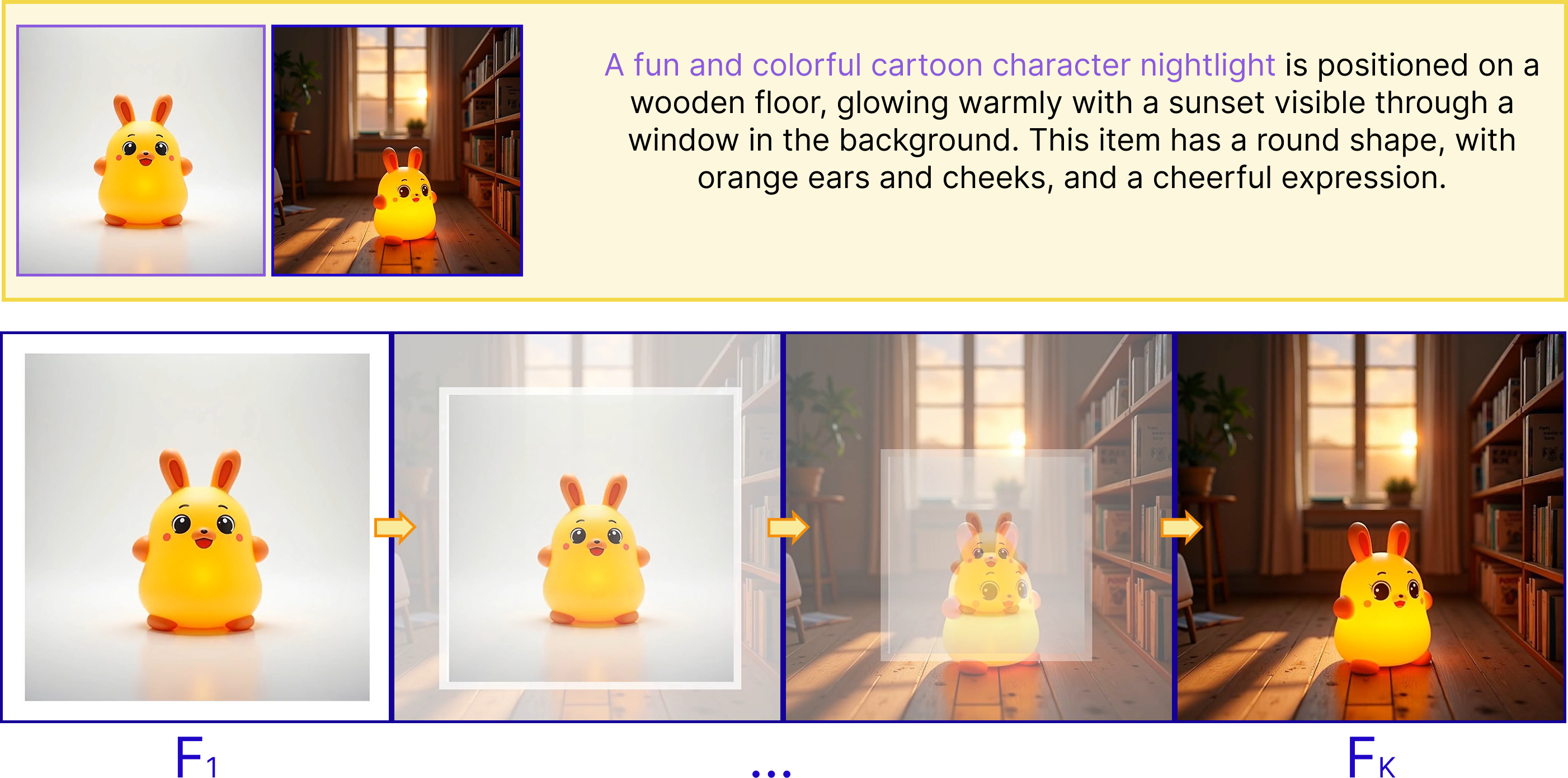}
    \caption{Training Data Generation Pipeline from a Subject-Driven Generation Paired Dataset. A short animation is built for transforming a white-background object image into an in-context scene described by the accompanying caption. Components in the yellow box are extracted from a Subject-200k subset \cite{chen2023subject}.}
    \label{fig:sup-subject}
\end{figure}

To preserve the text‑image alignment and reposing abilities of the pre‑trained I2V base model \cite{wang2025wan}, while encouraging plausible object placements (\eg, a laptop on a table, shoes on the floor), we use a filtered subset of the Subject‑200k dataset \cite{tan2024ominicontrol}. Each pair in this subset contains (a) a white‑background image of an object and (b) an image of the same object in a different context, together with a descriptive caption. We filter out pairs with object identity changes, often due to AI-generated images, using Grounding-DINO \cite{liu2024grounding}. We retain pairs where object descriptions yield bounding boxes with confidence $>0.55$ in both images. For each pair, we use the white-background image as $F_1$ and the contextual image as $F_K$ for generating short videos as in \cref{fig:sup-subject}. For objects undergoing significant reposing between the initial and final frames, we employ frame interpolation to create a smooth transition.


\subsection{Synthetic Side-by-side Compositions}

\begin{figure}
    \centering
    \includegraphics[width=\linewidth]{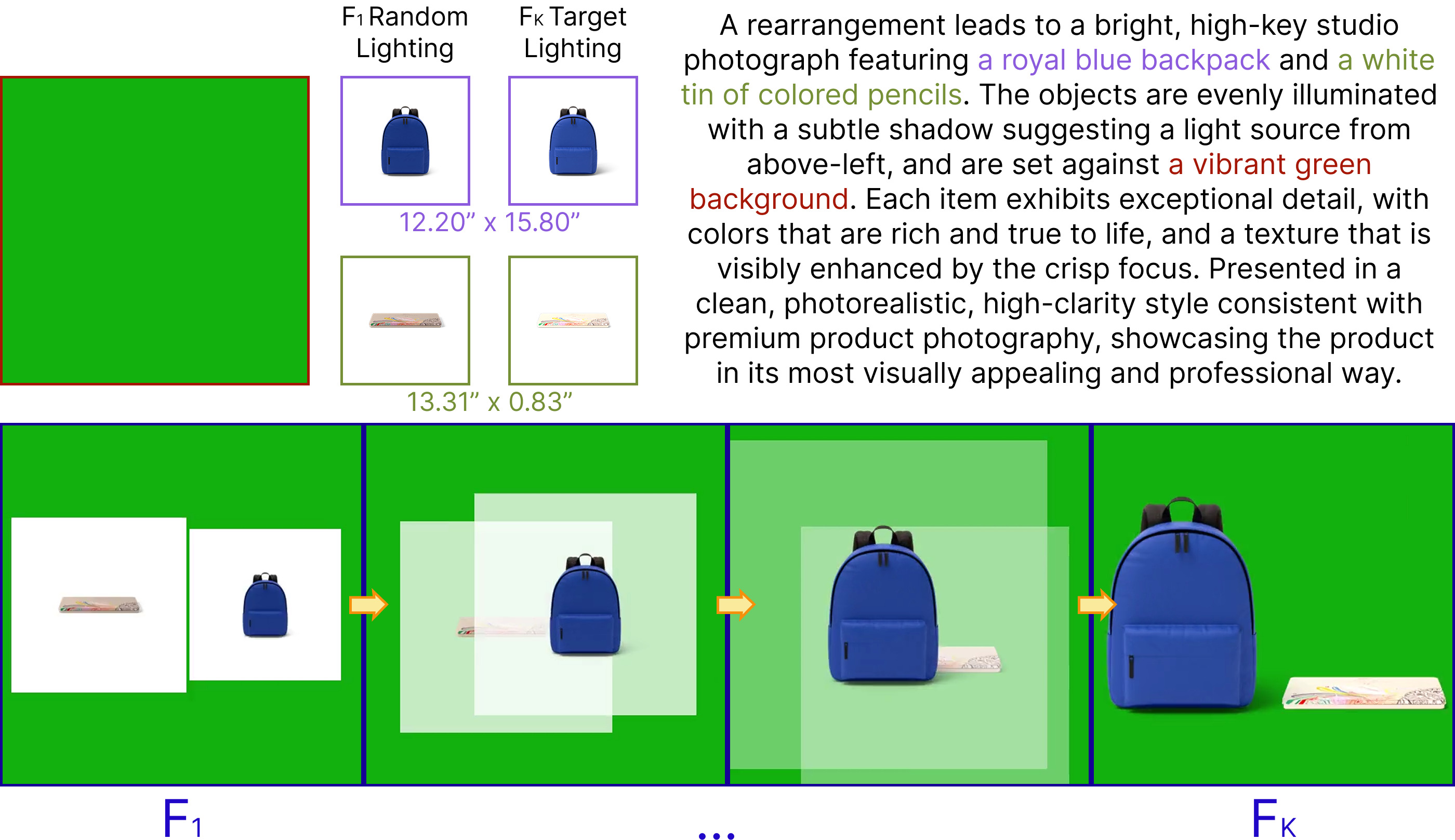}
    \caption{Training Data Generation Pipeline from Synthetic Side-by-side Compositions. The animation transforms a random arrangement of unsegmented objects with arbitrary scales and lighting on a plain background ($F_1$) into a coherent natural-looking scene ($F_K$). The final frame shows objects with consistent lighting and scales based on their real-life sizes. A template-based caption describes this transformation.}
    \label{fig:sup-RGBA}
\end{figure}

To enhance our model's ability to realistically rescale and relight objects, we augment our dataset with animations placing objects side-by-side on a shared ground in the final frame. We use 3D-rendered object images similar to those in \cite{collins2022abo}, clustering them into three size groups using K-Means \cite{mcqueen1967somekmeans}. For each video, we randomly sample two objects from one size group, extract their RGBA images with consistent lighting, and place them with shadows side-by-side on a random background for the final frame $F_K$. Importantly, when creating $F_K$, we use the known real-life sizes of the objects to scale them relative to each other, ensuring accurate size relationships. The initial frame $F_1$ is created by assembling images of the same objects with random lighting conditions. As illustrated in \cref{fig:sup-RGBA}, a short animation progressively relights, rescales, and repositions the objects from $F_1$ to $F_K$. The accompanying caption combines a template with object descriptions generated using \cite{bai2025qwen2}.

\subsection{Details on Data Augmentation}

\paragraph{Object Augmentation.} To discourage the model from simply copying the objects from the supplied images $I_{1..N}$, during training, each object is randomly (i) rescaled, (ii) rotated, and (iii) applied a small perspective warping. The resulting object images are scattered around the unchanged background in randomized order to minimize any rearrangement biases, while ensuring all objects are fully present in the image and there is no overlap between them. 

\paragraph{Background Augmentation.} To make the model resilient to a wide variety of background textures, each training sample using an aleatory background, draws it at random from one of three pools: (i) a collection of high‑quality photorealistic or textured background photos that match the aesthetic typically used by professional designers, (ii) plain backgrounds with a randomly sampled RGB color, and (iii) backgrounds created on the fly by combining simple primitives such as linear gradients, block textures, or radial gradients with harmonious color palettes. By mixing these three sources during training, the network learns to operate equally well on simple solid colors, patterned textures, and realistic photographic backdrops, thereby improving its robustness to any background it may encounter at test time.


\paragraph{Scene Completion.} During training we randomly apply a “scene‑completion’’ augmentation with a $20\%$ probability. In such cases the selected objects are treated as part of the background: their final positions are incorporated into the background description of the caption, and the background image (if one is provided) is edited to contain those objects throughout the entire K‑frame video. The remaining objects are still supplied as explicit conditioning inputs and are animated from their initial random locations to their target positions; in which they may even be partially occluded by the added background objects. This augmentation encourages the model to handle both added objects and items that naturally belong to the scene, improving robustness to varied compositional scenarios and diversity of generation. 

\paragraph{Design Elements.} With probability $10\%$ we treat a conditioning object as a design element rather than as an explicit visual cue. In that training sample, the object’s image $I_i$ is omitted from the visual conditioning set and does not appear in the initial frame of the video. Its description is inserted only in the caption $c$ (outside any $<OBJ> … </OBJ>$ block). Along the video duration, the object ``flies in'' from outside the image to reach its target pose in the final frame. This forces the model to rely on the textual description alone for synthesizing that item, encouraging robust text‑driven generation and improving editing skills. 

\paragraph{Object Replacement.} To enable the model to substitute objects in existing scenes, we introduce a replacement augmentation. When none of the objects in a training sample are designated as background elements or design‑only elements, with probability $7\%$, we select one object and substitute it with a new item in the generated animation. 
Simultaneously, the caption is updated to reflect the new edit. 

\section{Evaluation Dataset}
\label{sec:sup-testset}

As detailed in \refmainsec{4},
our evaluation set combines object images from the Amazon Berkeley Object Dataset (ABO) \cite{collins2022abo} and DreamBench++ \cite{peng2024dreambench}, with backgrounds from Unsplash \cite{unsplash} or plain-color canvases. The set comprises 122 combinations of objects, background images, and descriptive captions. Of these, 44 feature plain-colored backgrounds, while 78 use photorealistic images from Unsplash \cite{unsplash}. The object distribution varies: 17 sets contain a single object, 51 have two objects, 32 include three objects, 15 feature four objects, 6 contain five objects, and 1 set has seven objects. Descriptive captions were initially hand-written to ensure realistic and appealing compositions, then augmented via an LLM \cite{openai2025gptoss120bgptoss20bmodel} to incorporate diverse writing styles, varied compositions, and additional elements (\eg, the flower in \refmainfig{5} row 3). All figures in the main paper, except for \refmainfig{4} and \refmainfig{7}, as well as \cref{fig:sup-ablationdata,fig:sup-ablationtraining,fig:sup-soa1obj,fig:sup-soa2obj,fig:sup-soa3obj} in this SupMat, showcase inputs from this evaluation set. 

\section{Comparison to State of the Art}
\label{sec:sup-soa}

\begin{figure*}
    \centering
\includegraphics[width=0.675\linewidth]{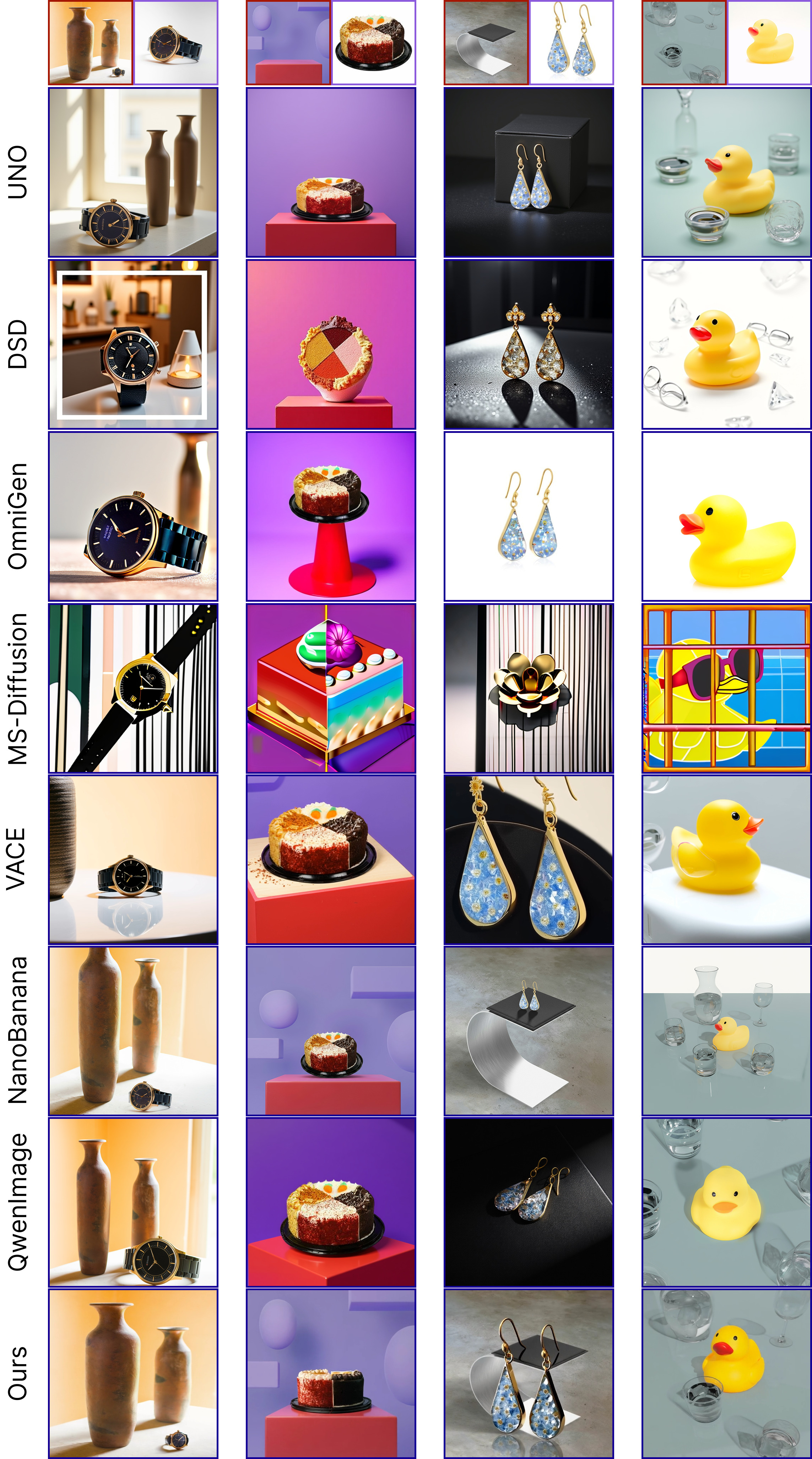}
    \caption{
    Comparison to State of the Art. Captions used to guide generation are, from left to right: (i) \textit{``A shiny \textcolor{obj1}{black and gold watch} is clearly displayed on the \textcolor{bg}{table next to the front vase and in front of the other}."}, (ii) \textit{``A \textcolor{obj1}{four flavours cake} is placed on top of a \textcolor{bg}{red platform, in front of a vibrant purple background}."}, (iii) \textit{``A \textcolor{obj1}{gorgeous pair of flowery drop golden earrings}  is showcased on \textcolor{bg}{the black surface in the image}."}, (iv) \textit{``A \textcolor{obj1}{yellow rubber duck} is carefully placed in the middle of the \textcolor{bg}{image and surrounded by see-through glasses}."} }
    \label{fig:sup-soa1obj}
\end{figure*}

\begin{figure*}
    \centering
    \includegraphics[width=0.675\linewidth]{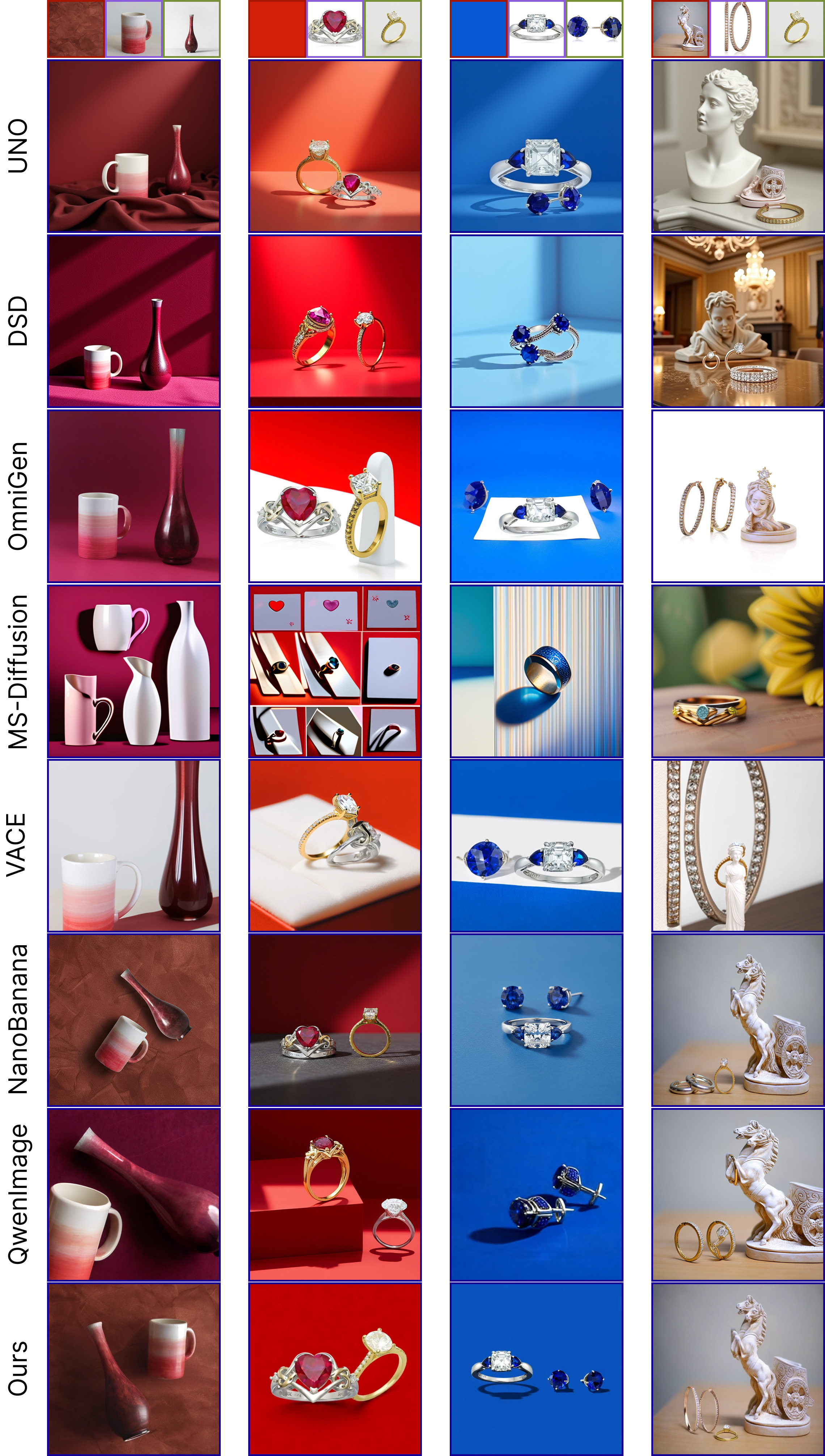}
    \caption{
    Comparison to State of the Art. Captions used to guide generation are, from left to right: (i) \textit{``A flat lay display of a \textcolor{obj1}{pink and white mug} and a \textcolor{obj2}{tall burgundy vase} laying down in a diagonal on a \textcolor{bg}{luxurious burgundy background}."}, (ii) \textit{``A \textcolor{obj1}{ring with a heart-shaped stone} is displayed next to a \textcolor{obj2}{classy ring with a big diamond} in a studio-style display with shadows using a \textcolor{bg}{red backdrop}."}, (iii) \textit{``A studio-style photo of a \textcolor{obj1}{silver ring with blue accents} and a \textcolor{obj2}{pair of blue earrings} on \textcolor{bg}{a blue backdrop}. The objects are close to the camera and casting a shadow to the side, laying on the same ground."}, (iv) \textit{``A \textcolor{obj1}{pair of hoop earrings} and a \textcolor{obj2}{diamond ring} are sitting on \textcolor{bg}{the table next to the statue}."} }
    \label{fig:sup-soa2obj}
\end{figure*}

\begin{figure*}
    \centering
    \includegraphics[width=0.675\linewidth]{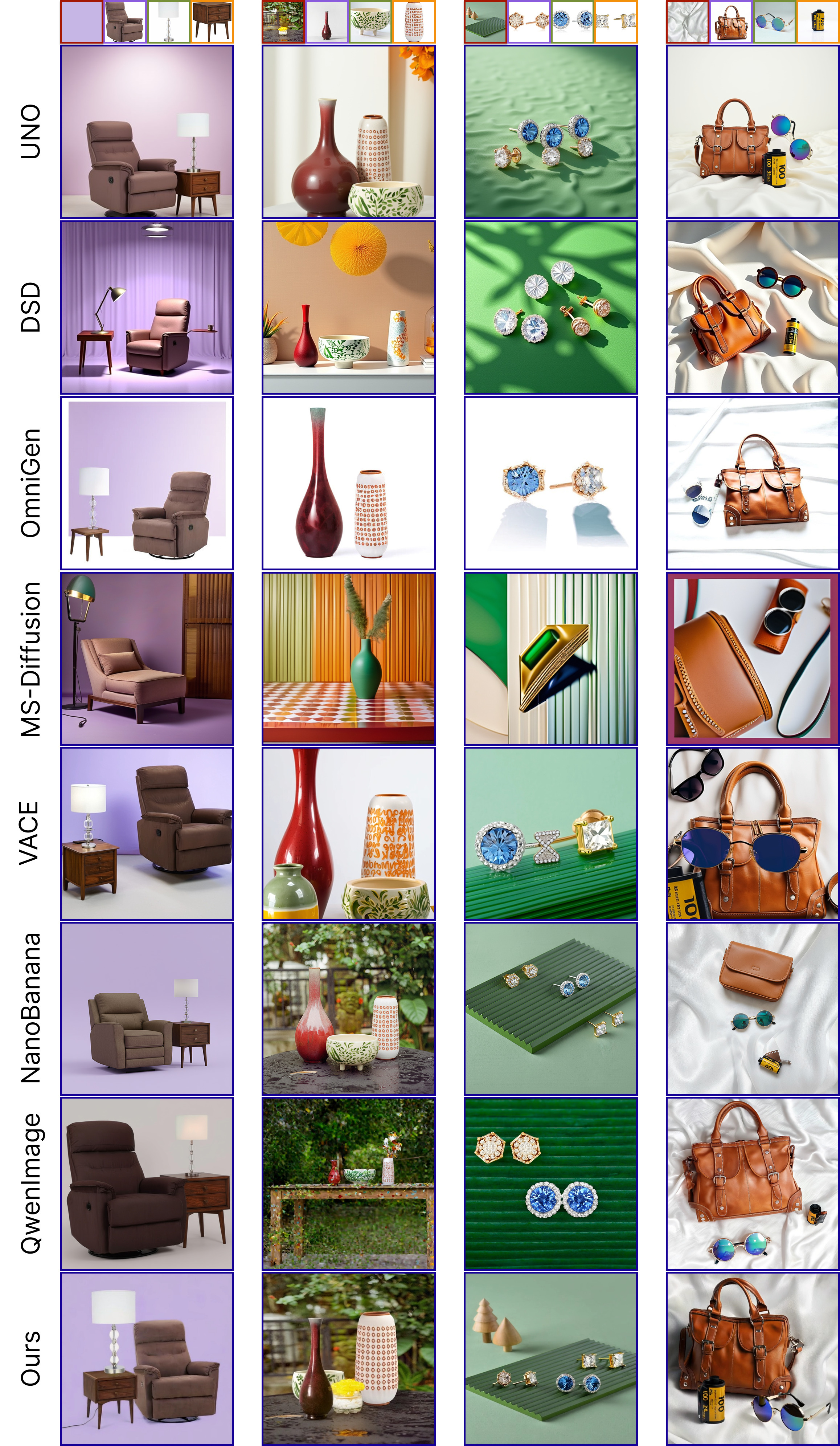}
    \caption{
    Comparison to State of the Art. Captions used to guide generation are, from left to right: (i) \textit{``A \textcolor{obj1}{brown recliner} , a \textcolor{obj2}{lamp} and a \textcolor{obj3}{wooden small table} are displayed in a studio-style image in front of a \textcolor{bg}{lilac backdrop}. A soft light from above illuminates the scene casting gentle shadows."}, (ii) \textit{``A \textcolor{obj1}{tall red vase}, a \textcolor{obj2}{white and green ceramics element} and a \textcolor{obj3}{tall white vase with orange elements} are clearly displayed on \textcolor{bg}{the table around and behind the white and yellow decoration}."}, (iii) \textit{``A \textcolor{obj1}{golden shiny pair of earrings}, \textcolor{obj2}{a silver and blue pair of earrings} and \textcolor{obj3}{a golden squared diamond earrings} are kept in a diagonal line on top of \textcolor{bg}{the wavy green pattern on the green backdrop}. They cast a soft long shadow to the bottom right of the image."}, (iv) \textit{``A \textcolor{obj1}{leather bag}, a \textcolor{obj2}{pair of retro sunglasses} and a \textcolor{obj3}{roll of film} are laying down on a \textcolor{bg}{satin white surface}."} }
    \label{fig:sup-soa3obj}
\end{figure*}

We show additional comparisons to state of the art models UNO \cite{wu2025uno}, DSD \cite{cai2025dsd}, OmniGen \cite{xiao2025omnigen}, MS-Diffusion \cite{wang2024ms}, VACE \cite{jiang2025vace}, NanoBanana \cite{comanici2025gemini}, and Qwen Image Edit \cite{wu2025qwen} in \cref{fig:sup-soa1obj} (one object compositing), \cref{fig:sup-soa2obj} (two object compositing), and \cref{fig:sup-soa3obj} (three object compositing). 

As shown in the main paper (\refmaintable{1}), some models such as OmniGen \cite{xiao2025omnigen} achieve higher identity preservation metrics than our model. However, this can be misleading, as illustrated in \cref{fig:sup-soa1obj} column 4. In this example, OmniGen's output obtains substantially higher identity preservation metrics (CLIP-I: 0.940, DINO: 0.869) compared to our model (CLIP-I: 0.855, DINO: 0.692). Paradoxically, our model better preserves fine-grained details, such as the duck's beak. This discrepancy arises because our model relights, harmonizes, and slightly reposes the object to integrate it realistically into the scene, while in this case OmniGen simply copy-pastes the object without any significant transformation or background integration. Consequently, while OmniGen maintains higher metric scores by not adapting the object to its new context, our model prioritizes realistic scene integration, which can slightly alter the object's appearance. The effectiveness of our approach is corroborated by the user studies presented in the main paper (\refmainfig{6}), where participants showed a preference for our model in terms of both identity preservation and overall image quality, despite the slightly lower metric scores. 


\section{Limitations}
\label{sec:sup-limitations}

\begin{figure}
    \centering
    \includegraphics[width=\linewidth]{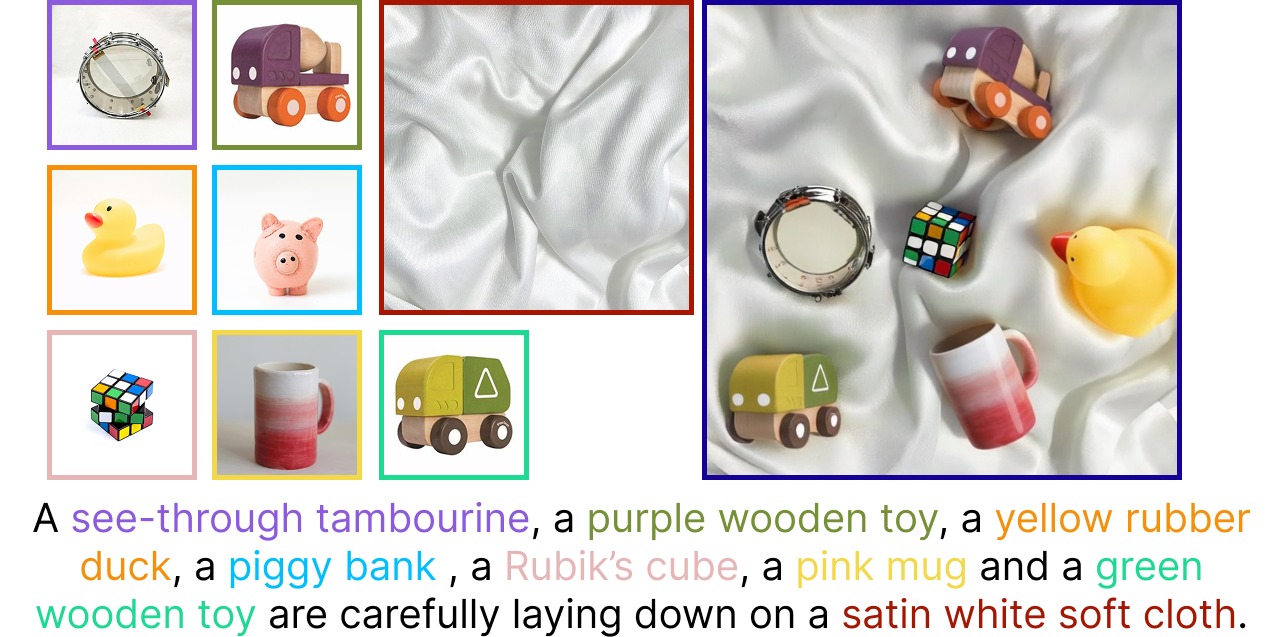}
    \caption{Model Limitation. Visualization of a failure case where the model struggles to introduce seven new objects into a scene, resulting in one object being omitted. }
    \label{fig:sup-limitation}
\end{figure}

We provide additional limitations of our model beyond those highlighted in \refmainsec{4.4}. First, since we use a video model to solve an image-to-image problem, the time and computational requirements at inference are slightly higher than if we used a similar architecture to generate a single image. However, we consider the benefits in terms of model versatility, image quality, completeness, identity and background preservation to outweigh this issue. If necessary, shorter videos could be used to train the model, reducing the gap between I2V and I2I models. Additionally, even though our model can successfully handle compositing a large number of objects in a scene, as shown in \refmainfig{1} (top, middle), the more inputs and constraints provided by the user, the more challenging the task becomes, resulting in an increased number of failure cases. We illustrate this in \cref{fig:sup-limitation} with an example where the model struggles to adhere to all constraints when attempting to composite seven objects into a natural, cohesive, and appealing display.

\end{document}